%% file: main.tex
\documentclass[twoside,onecolumn]{article}

\usepackage{blindtext} 
\usepackage{nameref}

\usepackage[sc]{mathpazo} 
\usepackage[T1]{fontenc} 
\usepackage{microtype} 

\usepackage[english]{babel} 

\usepackage[hmarginratio=1:1,top=32mm,columnsep=20pt]{geometry} 
\usepackage[hang, small,labelfont=bf,up,textfont=it,up]{caption} 
\usepackage{booktabs} 

\usepackage{lettrine} 

\usepackage{enumitem} 
\setlist[itemize]{noitemsep} 

\usepackage{abstract} 

\usepackage{titlesec} 
\titleformat{\section}[block]{\large\scshape}{\thesection.}{1em}{} 
\titleformat{\subsection}[block]{\large}{\thesubsection.}{1em}{} 

\usepackage{fancyhdr} 
\usepackage{titling} 

\usepackage{hyperref} 

\usepackage{tikz}
\usetikzlibrary{shapes.geometric, arrows, matrix, shadows, positioning}
\tikzstyle{process} = [rectangle, rounded corners, text width=7em, minimum height=1cm, text centered, draw=black, fill=white, drop shadow]
\tikzstyle{object} = [rectangle, text width=7em, minimum height=1cm, text centered, draw=black, fill=white, drop shadow]
\tikzstyle{arrow} = [thick,->,>=stealth]
\usepackage{url}

\usepackage{placeins} 

\usepackage{graphicx}

\usepackage[caption = false]{subfig}
\usepackage{float}

\usepackage[framemethod=tikz]{mdframed}
\usepackage{tabularx}

\usepackage{changepage}

\pretitle{\begin{center}\huge\bfseries} 
\posttitle{\end{center}} 
\title{A Base Camp for Scaling AI} 
\author{%
\textsc{C.J.C. Burges,~T. Hart,~Z. Yang,\thanks{Currently at Carnegie Mellon
    University. This work was done while Yang was an intern at MSR.}}
\and
\textsc{S. Cucerzan,~R.W. White,~A. Pastusiak,~J. Lewis} \\[1ex] 
\large Microsoft Research\\ 
}

\date{\today} 
\date{December 21st, 2016}
\renewcommand{\maketitlehookd}{%
\begin{abstract}
\input{abstract}
\end{abstract}
}

\addto\extrasenglish{%

}

\newcommand*{\fullref}[1]{\hyperref[{#1}]{\autoref*{#1} (\nameref*{#1}})}

\begin{document}
\maketitle
\newpage
\tableofcontents
\input{introduction}
\input{tal}
\input{fdl}
\input{discussion}
\input{ideas}
\input{ack}
\input{appendix_1}
\input{appendix_2}
\bibliographystyle{plain}
\bibliography{references}

\end{document}

%% file: abstract.tex
Modern statistical machine learning (SML) methods share a major limitation with the early approaches to AI: there is no scalable way to adapt them to new domains. Human learning solves this in part by leveraging a rich, shared, updateable world model. Such scalability requires modularity: updating part of the world model should not impact unrelated parts. We have argued that such modularity will require both ``correctability'' (so that errors can be corrected without introducing new errors) and ``interpretability'' (so that we can understand what components need correcting).

To achieve this, one could attempt to adapt state of the art SML systems to be interpretable and correctable; or one could see how far the simplest possible interpretable, correctable learning methods can take us, and try to control the limitations of SML methods by applying them only where needed. Here we focus on the latter approach and we investigate two main ideas: ``Teacher Assisted Learning'', which leverages crowd sourcing to learn language; and ``Factored Dialog Learning'', which factors the process of application development into roles where the language competencies needed are isolated, enabling non-experts to quickly create new applications.

We test these ideas in an ``Automated Personal Assistant'' (APA) setting, with two scenarios: that of detecting user intent from a user-APA dialog; and that of creating a class of event reminder applications, where a non-expert ``teacher'' can then create specific apps. For the intent detection task, we use a dataset of a thousand labeled utterances from user dialogs with Cortana, and we show that our approach matches state of the art SML methods, but in addition provides full transparency: the whole (editable) model can be summarized on one human-readable page. For the reminder app task, we ran small user studies to verify the efficacy of the approach.

%% file: introduction.tex
\section{Introduction and Related Work}
AI has had a long and difficult childhood \cite{Crevier:1993}, with nine ``AI Winters'' of
varying degrees commonly identified \cite{AIWinter:2016}. A key technical roadblock seems
to be the problem of scale, both in terms of tasks (it is much easier to build a system
that solves one domain-specific task well, than to build one whose domain can be easily
extended to new tasks) and complexity (it is easy to build small systems, but as they
grow, systems can rapidly become unmanageable). This has led to our reliance on
statistical methods, in particular on statistical machine learning (SML), where tasks are
essentially defined by labels on large datasets and by models that are carefully
constructed to solve those tasks. SML has had recent impressive successes
\cite{GooBenCou:2016}, and in fact major inroads have been made into the first three ``AI
Winters'' listed (``the failure of machine translation'', ``the abandonment of
connectionism'' and ``speech understanding research'') \cite{AIWinter:2016}.  But all
statistical machine learning methods to date share some critical limitations when it comes
to scalability: there is no procedure to make one machine-learned model maximally leverage
what another has learned, on a different task.  How can different models that solve
different problems be built in such a way that one can take advantage of the other's world
model, for shared basic concepts like space, time and number?  And how can one system take
advantage of the natural language learning that another has performed - how can we achieve
the desideratum: \textit{Every teacher benefits from what any teacher teaches}?  In the
world of SML, ``Transfer learning'' and its variants have been proposed to attack this
problem.  Transfer learning takes the form of models sharing features, parameters, or data
(for a review, see \cite{PanYan:2010}), but the set up is very close to the SML backbone;
the problem is viewed as a mismatch of the data distributions that one's model sees,
rather than the lack of a shared, fundamental, growable world model.

The key to a scalable world model is modularity.  We have argued that modularity in turn
requires \textit{correctability} \cite{Burges:2013}.  A system is correctable if its
errors can be corrected in such a way that its generalization performance is improved
while no (or, controllably few) new errors are introduced.  We have also argued that
\textit{interpretability}, by which we mean the ability of the designer to easily
understand why the system makes the assertions it does, is a desirable stepping stone
towards correctability, although perhaps not a necessary one.

The vast majority of current statistical models have complex, unseen, and essentially
uncontrollable decision surfaces that make full correctability and interpretability beyond
our reach.  (Statistical models unencumbered by such problems tend to be very small.)  In
these circumstances the simplest option available for error correction is to add labeled
training data and then either continue training starting from the current model, or
re-train. But this is a clumsy tool for a delicate task: previously correct predictions
can become errors, training data is expensive to gather, and we have little control over
whether the new model actually solves the errors in question.  One can certainly take a
more nuanced approach to balance robustness with improved generalization: see for example
(Wang et al., 2012) \cite{WanBenCol:2012} for an interesting application of such ideas to
ranking.  But currently, the tide is flowing in the opposite direction.  Large neural
network models are powerful but are becoming increasingly complex.  Furthermore, it is
worrying that deep nets for vision tasks generate their highest confidence outputs when
presented with inputs that to a human looks like noise \cite{NguYosClu:2014}; and that
almost imperceptible changes can be added to an input image that was previously classified
correctly by a deep neural net, causing that net to make an error (and that the same
changes, found using a sensitivity analysis for one net, can be applied to another,
trained on a subset of the same data, with the same result)
\cite{SzeZarSutBruErhGooFer:2013}.  These are striking departures from how we understand
the human visual system to work and would seem to be another barrier to both
interpretability and correctability.

The machine comprehension of text (MCT) presents particularly stark challenges for
statistical methods. An effective automated open domain dialog system will likely require
a rich world model and the ability to perform commonsense reasoning over it
\cite{Mueller:2014}. It is possible that purely statistical methods will be all that is
needed (see for example \cite{LiGalBroGaoDol:2016} for work on extending LSTM dialog
models to model individual speakers, and \cite{KimBanLi:2016} for using CNNs and RNNs for
dialog topic tracking).  But the performance of machines currently lags far behind that of
humans in comprehending even language familiar to a typical first or second grader.  Such
a dataset, in multiple choice format, was presented by \cite{RicBurRen:2013} and the state
of the art results on that data is approximately 70\% accuracy
\cite{WanBanGimMca:2015,TriYeYuaHeBacSul:2016}, compared to the 100\% that humans can
easily achieve.  A much larger question answering dataset provided by (Rajpurkar et al.,
2016) \cite{RajZhaLopLia:2016} for MCT paints a similar picture, with F1 scores achieved
by a strong machine learning baseline of 51\%, compared to 87\% for humans.

The vast majority of the approaches proposed not just for MCT, but for the general
algorithmic modeling of various aspects of human intelligence, are statistical in nature
(for recent work, see, for example, \cite{Liang:2016}, \cite{TerTenGer:2016},
\cite{PicMoo:2016} and \cite{HuYanSalXin:2016}).  This is a very appealing research
strategy for several well-known reasons.  But when it comes to building interpretable and
correctable methods, a natural question arises: how far can we get with methods for which
interpretability and correctability are built in from the ground up, before resorting to
statistical machine learning methods?  Specifically, can the problems associated with SML
be controlled by carefully compartmentalizing its role in such a system?

The work in this paper investigates this question.  We adopt a machine teaching approach,
and ask specifically: how much progress can be made with an interpretable, correctable,
rule-based system that can be taught using crowd-sourcing?  While the end goal is to
answer whether such a system could help address the core scalability issues facing AI
today, in this work we are simply searching for the hard limitations that the simplest
possible methods will run into; in particular, the points at which SML methods become
clearly required, if such points exist.  We view this exercise as constructing a base camp
for the expedition, and as we will see, the initial results are surprising: we find that a
crowd-sourced rule-based system, with generalization built in using a taxonomy, can
perform as well as a state of the art machine learning system, but correctably and
interpretably, on an MCT task.
    
Such a bottom-up approach is reminiscent of the early work in AI, perhaps the closest
being the work of (Schank, 1977), where particular scenarios are encapsulated by scripts
\cite{Schank:1977}.  The more recent work of (Hixon et al., 2015), in which a knowlege
graph is learned from conversational dialogs, is similar in spirit to ours: there, human
knowledge is leveraged more deeply than by just extracting labels, and an editable
ontology is used to aid in representing world knowledge.  However, their focus is very
different \cite{Hixon:2015}: they investigate the construction of a knowledge base to
support question answering, given a set of background true/false statements about the
world (in their case, 4th grade science), and they gather user input through dialogs
structured around multiple choice questions.  In contrast, our focus is on methods to
achieve scalability in general, on modeling language directly to support this, and on the
general development of apps for APAs, rather than question answering.

Our base camp is built on two foundations: {\em Teacher Assisted Learning (TAL)} and {\em
  Factored Dialog Learning (FDL)}, which we now describe.  Since FDL uses TAL but not vice
versa, we describe TAL first\footnote{We also use ``TAL'' to denote the overall dialog
  system itself, when the meaning is clear from context.}.

\subsection{Teacher Assisted Learning}
Traditionally, developing machine learning models has required expertise in machine
learning and programming.  However, some recent research has focused on how to help
someone who is a domain expert, but who knows little about machine learning, to quickly
build machine learned models
\cite{SimChiLakChaBotSuaGanAmeVerSuh:2014,WilNirDasLakSuaRedZwe:2015}.  LUIS
\cite{WilKamMokMilZwe:2015} provides user-friendly interfaces, available through a web
portal, that allow the user to perform necessary machine learning steps, such as defining
labels, adding examples, and training and evaluating models, in an interactive fashion.
LUIS also provides visualizations that show the prediction performance, and it integrates
an active learning component (``Suggest'') to allow the users to label the ``most
uncertain'' utterances.

However, a non-expert's knowledge and intuition about the problem domain is a highly
valuable resource that is not fully leveraged by traditional machine learning techniques,
beyond asking for labels.  For example, when predicting speaker intent for dialog
utterances \cite{Tur:2011}, a non-expert can not only easily interpret the meaning of each
utterance, but can also explain why they believe their interpretation, why some utterances
are ambiguous, when some utterances don't make sense, and so forth.  Current SML
approaches, including the above machine teaching methods, ignore this extra rich
information.

We hypothesize that a learning task can benefit from teachers who are experts in their
domain, but not in machine learning, in the two most important steps of learning:
representation learning and model generalization.  The core idea in TAL is to use
language-based templates and predicates which the teacher can tune to solve the problem at
hand, using examples from the training set and other resources for guidance.  TAL is by
construction interpretable and correctable.  In this setting, we have to address the
following fundamental questions:

\begin{itemize}
  \item TAL shares its objective with traditional machine learning, namely, maximizing
    generalization performance.  How can we make maximal use of the Teacher's language
    skills and domain expertise to achieve this?
  \item How can we build interpretability and correctability into the model from the
    ground up?
  \item As we assume the teachers are usually not technical experts, how do we
    design simple natural-language-based dialogs between TAL and the teacher
    during the teaching process, which will work for {\em any language modeling task},
    thus taking a step towards addressing the scale problem described above?
  \item Could this approach address the core scaling problem facing AI - that is, how can
    we ensure that all teachers (or systems) benefit from what any teacher teaches?
\end{itemize}

Because the representation is interpretable, highly modular, and language-based, the
Teacher can adjust it as desired, giving them much more control over how the system
detects meaning compared to SML approaches.  Teaching is an iterative process where, once
the teacher has taught a pattern, TAL can then run it over labeled data, looking for false
positives and negatives, which then guide the teacher to further tune their pattern.  This
approach thus leverages the deeper language skills mentioned above, because at any point
the teacher can reason why their pattern fails and take steps to correct it.

The work presented here focuses on the first three items above.  The fourth is a very
interesting open problem; some discussion is offered in Section \ref{sec:ideas}.

We present more details on TAL below.  In a case study in MCT, where the task is to detect
when the user has the intent to make a purchase, we demonstrate how an English speaker,
using a TAL instance that integrates the WordNet semantic taxonomy \cite{Fellbaum:1998},
can train an intent prediction model for dialog utterances, correctably and interpretably,
and attain results comparable with LUIS \cite{WilKamMokMilZwe:2015}.

\subsection{Factored Dialog Learning}
\emph{Automated Personal Assistants} (APAs) have a long history in AI (see for example
\cite{Maes:1994,Myers:2007}) and modern APAs, such as Siri or Cortana, can perform various
useful tasks through conversational dialogs with users: for example, they can detect when
the user wants to make a phone call, or find an address, or add an item to their calendar,
and can then take an appropriate action.  However, scaling APAs to large numbers of
different kinds of tasks, and eventually, to having them anticipate the user's needs
through natural, engaging dialog, is still beyond us.  The traditional software
development process \cite{Boehm:1988} is not well suited to this problem.  Typically, to
support a new task, designers and developers must first write and review the new design
and its requirements, they must make necessary changes to the implementation (if not start
again from scratch), and finally, the new system must undergo a suite of unit and system
wide tests, which may require beta testing with users.  For dialog apps, these changes
include redesigning the turn-taking structure of the dialog and implementing any extra
needed underlying logic.  Moreover, the language skills (for both speaking and
understanding) of an APA are crucial for a good user experience, from intent detection to
task completion.  This process thus usually requires the developers to ``teach'' the
languages the agent needs for each application, in some form, and to continue training to
understand what the user says even if the domain is fixed.

Here we investigate addressing this scalability issue by leveraging two key ideas: first,
by factoring the development in such a way that the entire language learning part of the
problem can be performed by non-experts; and second, by developing a shared world model
that can be re-used by widely different classes of applications.  The goal is to achieve
scalability by opening up the possibility of using crowd sourcing for application
development (in addition to using crowd source for language learning, as above).  We call
this approach \textit{Factored Dialog Learning} (FDL).

FDL uses TAL as a language learning component, but it also introduces two additional roles
to those found in the traditional development process: the \emph{Teacher} and the
\emph{Designer}, each of which deals with a different aspect of the scalability issue.  We
define the Teacher as the person who teaches the APA the language skills required to
engage in a user dialog, for a particular application.  We define the Designer as the
person who identifies a {\em class} of tasks that they want the APA to handle, and who
then designs the abstract dialog flow for every application domain within that class
(including user-system dialog shared across all applications in the class).

As opposed to the complex skillset required of the developer in the traditional
development process, here, the Teacher need only be familiar with the application that
they want to build, in addition to being fluent in their native language.  The FDL
framework requires more of the Designer (but still far less expertise than that required
of a developer): the Designer must learn a simple, high level scripting language which we
call the Module Specification Language (MSL).  The Designer uses the MSL to specify how
the Teacher will create an app from the Designer's class of applications.  Although MSL is
a simple scripting language, it does support basic programming concepts such as loops and
conditionals.  We will describe the MSL in detail below.

In this paper, in order to contrast the different roles played by FDL and TAL, we choose a
development example where FDL's dependence on TAL is very light: we implement FDL for a
class of {\em Event Reminder} applications, and we demonstrate its use to create reminder
scheduling apps for tracking the consumption of medications, visits to the gym, birthdays,
and some other reminder tasks.

FDL factors the language development out of the overall development process and thus can
assign all language development tasks to the Designer and Teacher.  However, of course the
Developer still plays a crucial role in the overall process.  For example, for the Event
Reminder class, there are seven quantities that the APA must extract from the user, given
that the user wants to create a reminder: two strings (a name for the reminder itself, and
any notes to add to the reminder) and five numbers (the event duration, frequency, group
frequency, start time, and end time).  These will be explained more below, but the point
here is that the developer must write the code to have the APA do the appropriate thing
once the user is running the app and this data is received (in this case, the APA would
simply add the notated events to the calendar).  However, notice that the APA's actions
triggered by these seven quantities are independent of the particular reminder app being
run.  Thus it is the Developer's (and Designer's) role to identify \textit{classes} of
applications such that the APA actions taken for any app in a given class need only be
coded once.

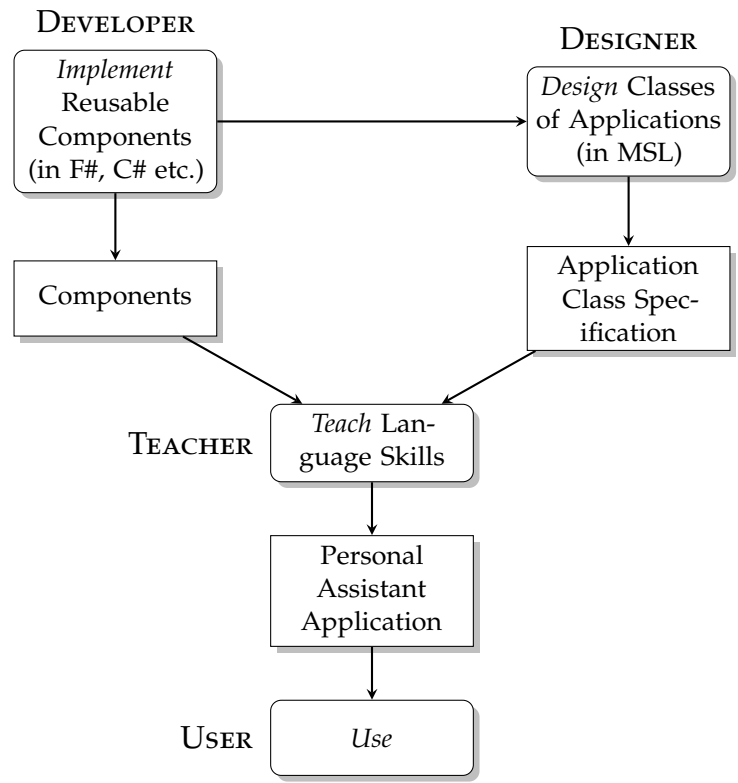
\begin{figure*}[h!]
  \centering
  \begin{tikzpicture}
    \matrix [row sep=2em, column sep=2em] {
      \node (implement) [process] {{\em Implement} Reusable Components (in F\#, C\# etc.)}; & & \node (design) [process] {{\em Design} Classes of Applications (in MSL)}; \\ 
      \node (component) [object] {Components}; & & \node (class) [object] {Application Class Specification}; \\
      & \node (teach) [process] {{\em Teach} Language Skills}; & \\
      & \node (app) [object] {Personal Assistant Application}; & \\
      & \node (use) [process] {\em Use}; & \\
    };
    \draw [arrow] (implement) -- (component);
    \draw [arrow] (implement) -- (design);
    \draw [arrow] (component) -- (teach);
    \draw [arrow] (design) -- (class);
    \draw [arrow] (class) -- (teach);
    \draw [arrow] (teach) -- (app);
    \draw [arrow] (app) -- (use);
    \node[above = 0.3em of implement] {\textsc {\large Developer}};
    \node[above = 0.3em of design] {\textsc {\large Designer}};
    \node[left = 0.3em of teach] {\textsc {\large Teacher}};
    \node[left = 0.3em of use] {\textsc {\large User}};
  \end{tikzpicture}
  \caption{The Factored Dialog Learning (FDL) framework.}
  \label{fig:diagram}
\end{figure*}

Figure \ref{fig:diagram} summarizes the development process flow diagram of the FDL
framework.

\subsection{Predicates, Templates, and TAL and FDL compared}
The key concept underlying both TAL and FDL is that of a {\em predicate}.  A predicate can
be thought of as a Boolean function over text where the function has 'state'.  The state
is modeled with a data structure that contains slots and handler functions.  When the
predicate fires (i.e. takes the value 'true'), its slots are populated, and its handler
functions tell the APA what to do.

Thus for example, a predicate for the concept \textit{is collocated with} might contain a
slot for the set of entities that are deemed to be collocated, whenever the predicate
returns true.  No constraints are imposed on the text: in particular, we do
not require the presence of a verb\footnote{Our earlier work did, and predicates are
  associated with verbs in theories of syntax and grammar
  \cite{PredicateGrammar:2016}. However, queries submitted to intelligent agents often
  lack verbs but nonetheless convey intent(for example, self-reminders like {\em Sophie's
    piano lesson}).}.

TAL distinguishes two kinds of predicates: \textit{Parsed Predicates} and
\textit{Learned Predicates}.  Parsed predicates currently form TAL's world model; they
encapsulate notions such as time, space, and number, and they are currently hardwired.
Parsed predicates are thus easily shared across classes of applications.  For example,
in building the event reminder class of applications described in this paper, the
developer had to create parsed predicates for time, frequency, period, and duration -
which could then all be used by any other designer in their creation of a different class
of applications.  It seems reasonable to allow that the world model, at least at the level
of physics and mathematics, warrants its own special development effort, since physics and
mathematics themselves form such a succinct yet powerful world model\footnote{The only
alternative that comes to mind is to try to learn the laws that govern the physical
world from labeled examples. But at its core, SML is for modeling functions that we
don't know how to describe in closed mathematical form, which is not the case here.}.

Learned predicates, on the other hand, are used to model all other forms of meaning in
language.  We thus expect that learned predicates will greatly outnumber parsed predicates
as TAL grows.  Learned predicates are not hard-coded but instead are learned entirely
using human teachers.  For learned predicates, we divide the problem, and introduce {\em
  templates}, which are used to detect patterns in language.  A learned predicate is
declared to be true if and only if its containing template ``fires'' (matches a text
fragment).  Templates are also entirely taught (rather than coded): they are
language-based and they leverage a semantic taxonomy, as well as teacher input, to achieve
generalization.  In this study we use WordNet for our semantic taxonomy
\cite{Fellbaum:1998}, but teachers can also both add and remove items from the taxonomy as
needed.

Thus, any given template has an associated set of learned predicates, and whenever that
template matches a piece of text, its predicates are declared true for that text.  For
example, the text \textit{``Jesse ate a sandwich''} might fire a ``Collocated'' predicate,
and also a ``Consumes'' predicate.  In addition, a given predicate can be fired by several
different templates: for example, \textit{``Jesse ate a sandwich''} and \textit{``The sponge
  absorbed the liquid''} might both fire a ``One entity consumes another'' predicate, even
though the templates that fire that predicate are different.

Since learned predicates and their containing templates are all learned from teachers, TAL
attacks the scalability problem by leveraging the expertise of crowd-sourced teachers who
need only be experts in their own native language.  By factoring the problem in this way,
the pool of available teachers becomes very large.

In contrast to TAL, FDL's main task is to factor the language needed for the application
development process in such a way that one designer can create a script that could be used
by hundreds of teachers, who then in turn create apps to support millions of users.  Thus
FDL attacks the scalability problem by further factoring the language modeling problem in
such a way that the available expertise matches each subtask well.

In our Event Reminder class of apps described below, the TAL system relies almost entirely
on parsed predicates, and it had to learn only one extra learned predicate (to allow it to
detect when the user wants to call up the app).  (When TAL searches for predicates
matching a piece of text, it always tests both parsed and learned predicates.)  However in
our study of learning user intent, TAL relies entirely on learned predicates.  We chose to
use these two examples with as little overlap of the predicate types they need as
possible, to more clearly test, and contrast, the ideas in each approach.

%% file: tal.tex
\section{Teacher Assisted Learning}
\label{sec:TAL}
In the following, for succinctness, we replace the phrase ``correctable and
interpretable'', as defined above, with ``transparent''.

The main design goal for Teacher Assisted Learning (TAL) is that the resulting models be
fully transparent: TAL's behaviour should always be fully and explicitly attributable to
the underlying data and rules (the WordNet taxonomy, our gazetteers, the hardwired lists
of tokens (e.g. personal pronouns)), or to the data and rules input by a Teacher.  In
earlier versions of the system we used dependency and consituency tree parsing, semantic
role labeling, and other NLP constructs, using the SPLAT NLP tool from MSR
\cite{QuiChoGaoSuzTouGamYihVanChe12} and the Stanford NLP package
\cite{ManSurBauFinBetMcc:2014}.  These tools are the culmination of decades of development
effort and it seemed wise to use them rather than try to reinvent the wheel.  But using
them revealed four significant drawbacks: they are brittle, they are slow, they are not
transparent, and they solve a harder problem than we need to solve.  The brittleness and
lack of transparency are unavoidable consequences of the statistical models underlying the
tools; one could investigate building ``shims'' around the tools to correct errors, and
then retrain the underlying models when enough data was gathered, but the overhead would
be high, and retraining does not guarantee that the errors are resolved.  The speed issues
are harder to address and would likely have proven a show stopper to shipping TAL (imagine
having to compute the parse trees and SRL for the inputs from millions of users,
simultaneously).  At the time, we only had a hunch that these systems also solve a harder
problem than we really needed to solve, but this indeed turned out to be the case.  In
this section, we describe in detail the current processing that TAL uses.

\subsection{Design}
Here we extend the high level overview given in the Introduction to give a detailed
description of templates and predicates, and of the process used to train TAL.

\subsubsection{Templates} \label{sec:Templates}
\label{sec:templates}
The template data structure consists of an ordered list of items used for pattern
matching, together with a pointer to a set of predicates.  Each item can be roughly
thought of as either a string, a synset, or a set of synsets (detais are given below).
For efficiency, templates are indexed by the first synset occuring in their list (every
template must have at least one synset); thus, rather than scanning all templates over a
piece of text to detect patterns, only those templates indexed by the possible synsets
correspoding to the current token in the text are tested for possible matches with that
and subsequent tokens.  A template is said to ``match'' or ``fire'' if and only if a
sequence of tokens in the text is found that matches that template's list, in order.
During the matching process, only tokens that could be nouns, verbs, adjectives, adverbs,
pronouns, or negation terms, are considered.  If a template fires, then TAL asserts that
all of the predicates associated with that template are true, for that piece of text.  As
mentioned above, the template/predicate map is many-to-many: a given predicate can also be
pointed at by more than one template.

Each template's list is of fixed length but different templates can have different length
lists.  There is no limit to the possible length of a template's list, but typically the
list has between two and four elements: see Table \ref{tab:histogram} for the histogram of
the lengths of list for the TAL classifier trained for this paper (see Section 
\ref{sec:classifier} for details).

\begin{table}[!ht]
\centering
\begin{tabular}{cc}
\toprule
List Length & Number of Lists \\
\cmidrule(r){1-1} \cmidrule(l){2-2}
2 & 1 \\
3 & 8 \\
4 & 14 \\
5 & 1 \\
\bottomrule
\end{tabular}
\caption{Histogram of template list lengths for the TAL classifier trained for this paper.}
\label{tab:histogram}
\end{table}

Concretely, each item in a template's list can be one of three types: a string; a set of
noun synsets; or a pair, consisting of a single verb synset, together with a simplified
form of that verb's tense.  Currently TAL supports the simple tense types {\em Past}, {\em
  Present or Future}, or the catchall {\em PastOrPresentOrFuture}.  An element in the
template matches a token in the text if one of the following three conditions holds:

\begin{enumerate}
\item The element is a string, and the token is the same string.
\item The element is a single verb synset $S$ with simple tense $T$, and the token appears
  as a possible verb in the taxonomy, such that one of the possible synsets of that verb
  is a hyponym of $S$, and such that that token could also have simple tense
  $T$\footnote{Note that a single token can have several possible simple tenses: for
    example ``cut'' is the infinitive/present, past tense, and past participle of the verb
    ``to cut''.}.
\item The element is a set of noun synsets $\mathcal S$, and the token appears as a possible
  noun in the taxonomy, such that one of the possible synsets of that noun is a hyponym of
  one or more of the synsets appearing in $\mathcal S$.
\end{enumerate}

We reserve the use of string matching for parts of speech that have a small number of
possible instantiations: currently, TAL only allows question adverbs, as listed in Table
\ref{tab:questionAdverbs}.  Note that if needed, other specialized parts of speech
(e.g. prepositions) could also be used (for example, {\em ``Let's get Taco Bell''} and
    {\em ``Let's get to Taco Bell''} connote different user intents).  We decided to make
    question adverbs the first such direct match type, to give TAL rudimentary question
    detection abilities.

\begin{table}[!ht]
\centering
\begin{tabular}{cccccc}
\toprule
who &  whoever & when & whenever &  what & whatever \\
how & however & why & whyever & where & wherever \\
\bottomrule
\end{tabular}
\caption{Question adverb strings used by TAL for direct matching.}
\label{tab:questionAdverbs}
\end{table}

During processing, template matching is not done over the raw text, but over a list of
{\em parsed components}.  The construction of these is described below.  Here we just note
that while our templates currently model nouns, verbs and question adverbs, our parsed
components in addition model adjectives, general adverbs, prepositions, pronouns, and
negation.  Some of these take part in the current matching process (see below), while the
others were included to facilitate the extension of the template data structure if
desired.

\subsubsection{Predicates}
As mentioned in the Introduction, TAL distinguished two kinds of predicates: {\em learned
  predicates}, which a Teacher constructs by interacting with TAL via natural language
only, and {\em parsed predicates}, which form the backbone of TAL's world model and
which are currently hardwired.

{\underline {\textbf Learned Predicates:}} The learned predicate data structure consists
of the predicate's name (a string), the names of its components (a list of strings), a
slot for optional notes (also a string), and a set of handlers for specifying the scripts
and actions to be launched when this predicate fires (which occurs when and only when any
template that points to it fires).  For example, a predicate modeling ``Entity ingests
something'' might have component names ``Ingester'' and ``Ingested''.  It is important to
note that mapping the text to these component names is straightforward, since TAL can just
ask the Teacher for the mapping once the template has been trained.  Thus {\em ``It was
  eaten by him''} would fire a different template than that fired by {\em ``He ate it''}
(recall that templates take word order into account) and the slots, at run time, would be
filled automatically, based on the mapping specified by the Teacher at train time.  This
is an important advantage of using Teachers: we do not have to solve the Semantic Role
Labeling problem to compute this mapping. Table \ref{tab:LearnedPredProps} shows the
properties of the \textbf{LearnedPredicate}.  The last item - LearnedSynsetMap - was added
purely to lower the cognitive load on the Teacher: the chances are good that a synset
found for a word used to train a previous templates for a given predicate, is also the
correct synset for that word, for a different template being trained for that same
predicate.  (The Teacher always has the option of choosing a different synset.)

\FloatBarrier
\begin{table}[h!]
\centering
\begin{mdframed}
\begin{tabularx}{\textwidth}{l|X}
  Name & Description \\
  \hline
  \hline
  PredicateName & The name of this predicate. \\
  \hline
  ComponentNames & The mapping between the matching template and the relevant item of the
predicate, such as a value that would be retrieved by an application that receives notice
the predicate has fired. \\
  \hline
  Notes & Additional information about the predicate. \\
  \hline
  Handlers & A set of action names that map to the script that is to be executed to
implement that action when the predicate has fired. See \fullref{sec:Handlers} for a
detailed description. \\
  \hline
  LearnedSynsetMap & A mapping from a word to the last synset that was selected for that
word for this predicate. Used to eliminate duplicate questions to the Teacher during
predicate training, and presents a confirmation to use this previous selection, which the
Teacher may override. \\
\end{tabularx}
\end{mdframed}
\caption{LearnedPredicate properties.}
\label{tab:LearnedPredProps}
\end{table}
\FloatBarrier

{\underline {\textbf Parsed Predicates:}} So far we have written built-in parsed
predicates to support notions of time (for the Event Reminder module described below).
A parsed predicate has a name and may have properties. Table \ref{tab:BIPPsInBrief} gives a
brief description of the built-in \textbf{ParsedPredicates} currently implemented by FDL;
see table \ref{tab:BIPPs} in \fullref{turn:TestPredicateTurn} for a detailed description
for script usage.

\FloatBarrier
\begin{table}[h!]
\centering
\begin{mdframed}
\begin{tabularx}{\textwidth}{l|X}
  Name & Description \\
  \hline
  \hline
  Frequency & This has properties for the period of time (e.g. ``day''), the number of
events per period (which can either be single integer, or a range of integers, as in
``three to four times per day''), and the starting time of the first occurrence in each
period. \\
  \hline
  SequenceDuration & The duration of the sequence of reminders (how long the reminders
are to remain on the user's calendar; e.g. the number of recurring events to place in the
user's calendar). \\
  \hline
  EventDuration & The duration of a single event (e.g. a medication reminder might specify
5 minutes, while an exercise reminder might use two hours). \\
  \hline
  StartDate & The date the sequence of reminders is to start (e.g. a specific date or
``tomorrow''). \\
\end{tabularx}
\end{mdframed}
\caption{TAL's current built-in Parsed Predicates.}
\label{tab:BIPPsInBrief}
\end{table}
\FloatBarrier

Parsed predicates do not use templates; they are fired by (currently built-in) pattern
detectors. We chose to hard-wire several pattern detectors since these patterns are likely
to be shared across many different applications, and since the quantities they represent
can be modeled extremely succinctly by mathematical notions such as time, and the number
line.  We wrote patterns to model date, time, duration and frequency.  As an example,
Table \ref{tab:frequency} shows the pattern detectors used to detect frequencies from
English text.  Each pattern detector is written as an F\# {\em active pattern}; we found
the F\# language to be particularly well-suited for this kind of modeling.  Each active
pattern's name encapsulates one of the possible patterns it captures, to facilitate code
readability: thus for example, the {\em OncePerSecond} active pattern models text snippets
such as {\em ``Once a second''}, {\em ``Twice per day''} or {\em ``Three times every
  week''}.  The order of the scans (from top to bottom, in Table \ref{tab:frequency})
matters: more complex patterns can contain simpler ones (e.g. the ``Hourly'' detector
should be scanned over the text only if the ``OnceHourly'' detector failed to fire).

\begin{table}[!ht]
\centering
\begin{tabular}{c}
\toprule
OncePerSecond \\
OnceHourly \\
NtoMTimesHourly \\
NtoMTimesASecond \\
OnceEveryNtoMSeconds \\
NtoMTimesEveryNtoMSeconds \\
EveryNtoMSeconds \\
EverySecond \\
Hourly \\
NtoMHertz \\
\bottomrule
\end{tabular}
\caption{The scan used to detect frquencies in text.  Pattern detectors are run in order,
  from first to last, above.}
\label{tab:frequency}
\end{table}

\subsubsection{Text Preprocessing}
\label{sec:talPreprocessing}
Since we use no more sophisticated standard NLP processing than lemmatization and basic
part of speech tagging, we can describe the process in full here.

First, the user's input is broken into sentences.  Then, terminators (characters in the
set \{``.'', ``!'', ``?''\}) are removed from the end of each sentence, and any leading or
trailing space is then removed\footnote{Currently TAL only handles single sentence input;
  if it detects more than one sentence at this point it issues a warning and takes only
  the first.}.  The text is lower-cased, and then any tokens that appear in the Slang Map
(see Appendix \ref{apx:slangMap}) are replaced (for example, ``wanna'' is replaced by
``want to''). Apostrophes (``'s'' and ``''') are then removed.  Then two mappings
corresponding to TAL's internal model of time intervals are performed: first, strings
matching the pattern {\em \#-\#}, where {\em \#} represent an integer (i.e. dashed
intervals), are mapped to {\em \# to \#}.  Then, strings matching the pattern {\em \#x/T},
where ``\#'' is an integer and {\em T} is one of the hardwired time intervals (e.g. {\em
  day}) (i.e. slashed intervals) are expanded to the form {\em \# times per T}.  We expect
these mappings to be shared across all applications that use TAL's world model and so they
are hardwired.  Finally, tokens are mapped to lemmas as follows: we use two files, one
that lists verbs and their declinations, and one that lists nouns and their plurals.  If a
token appears as a declination of a verb, and that declination does not appear in the noun
list, then that token is declared to have type verb. Otherwise, we gather all possible
parts of speech for the token. In all cases, tokens are replaced by their lemmas, and for
verbs, their declination is mapped to one of {\em Past, PresentOrFuture,
PastPresentOrFuture}.  Throughout processing, both the original tokens and their lemmas
are kept.

\subsubsection{The Parsed Components List}
We then construct a {\em parsed components list} for the sentence.  This amounts to
mapping each token, or compound phrase, to a set of its possible parts of speech (POS);
keeping track of negations; and handling modal verbs.  The parts of speech we track are
shown in Table \ref{tab:pos}.  We reserve the 'Unkown' flag to model tokens whose POS does
not map to one of those listed.  Note that the code that implements the processing
described in this section is language-specific and currently would have to be rewritten to
map TAL to handle new languages, although the amount of code is small.

\begin{table}[!ht]
\centering
\begin{tabular}{c}
\toprule
Noun \\
Verb \\
Adverb \\
Adjective \\
Preposition \\
Personal Pronoun \\
Non-personal Pronoun \\
Unknown \\
\bottomrule
\end{tabular}
\caption{Tracked parts of speech.}
\label{tab:pos}
\end{table}

{\bf Compound Phrases:} In Teach phase, known compound phrases (phrases consisting of two
or more tokens that appear in our extended WordNet taxonomy, for example {\em look for} or
{\em White House}) are shown to the Teacher for confirmation, and the Teacher can then
either confirm, or declare a found compound phrase as non-compound (for example, {\em get
  to} has three synsets in WordNet, but none of them have the meaning as in {\em I need to
  get to the store}).  Then the Teacher is given the option of adding any compound phrases
that TAL missed (for example, the name of a new video game).  New compound phrases thus
formed are then added to the taxonomy.  In Test phase, currently, TAL assumes that any
compound phrases it finds are correct (but see Section \ref{sec:ideas}).

{\bf Negation:} We simply track negation, as the negation of the tracked part of speech
immediately following the negator.  It turns out that every part of speech we track can be
negated; Table \ref{tab:negation} gives an example for each.

\begin{table}[!ht]
\centering
\begin{tabular}{c}
\toprule
Noun: "No deal!" \\
Verb: "I did not walk." \\
Adverb: "I walked, but not quickly." \\
Adjective: "The paint is not wet." \\
Preposition: "This is not for me." \\
PersonalPronoun: "It was not me." \\
NonPersonalPronoun: "That is not it." \\
\bottomrule
\end{tabular}
\caption{Negation examples.}
\label{tab:negation}
\end{table}

{\bf Modal Verbs:} We distinguish {\em semi-modal} verbs from {\em pure modal} ones.  A
modal is ``semi'' if it can be a standalone verb, or a modal verb, such as ``have'' or
``did''.  A modal is ``pure'' if it cannot occur as a standalone verb.  Our list of pure
modals is {\em can; could; may; might; must; shall; should; will; would}.  We include
modals like {\em can} and {\em must} since although they can occur in their own sentence
(e.g. {\em I can.}), they always refer to an action.  We also cast modals as past tense,
or 'present or future' tense, and update the verb's tense based on its modal, if it has
one.  Finally we also distinguish modals with possible noun meanings ({\em can; may; will;
  might; must}).

Processing of modals occurs as follows.  As mentioned, negation is detected and attached
to the following POS, and verb tense is detected attached to a following verb.  A pure
modal that can also be a noun is declared a noun if it is preceded by a determiner or a
possessive pronoun (e.g. {\em The can is heavy.}).  Otherwise, pure modals themselves are
not kept for further processing.  For semi modals, if they modify a verb, that verb's
tense is kept, and the modal is skipped; if not, they are kept as 'stand alone' verbs
(e.g. {\em I have one.}).

Finally, all parsed components that could not {\em possibly} be a noun, verb, pronoun, or
directly matchable token (which in the current system are restricted to question adverbs)
are dropped; the remaining noun and verb meanings are mapped to sets of possible synsets
(in Teach phase, the Teacher is first asked to identify whether the token is a noun, verb
or neither, if it's ambiguous); and certain pronouns are mapped to special purpose synsets
(for example, the pronoun {\em I} is mapped to a synset which derives from the first noun
meaning of {\em narrator}); and personal pronouns other than {\em I, you} or {\em we} are
mapped to the synset {\em person.n.01}).  This was done mainly to make patterns containing
these pronouns more clearly readable.

\subsubsection{The Template Matching Process}
\label{sec:TemplateMatching}
Template matching to a phrase is straightforward.  First, the phrase is mapped to its
parsed component list as above.  The template matching is done by an ordered comparison
between the template's list, and the phrase's parsed component list.  To greatly reduce
computational overhead, templates are stored in a dictionary, where the key is also the
first synset occurring in the template's list.  In order to search for a match at a given
position in the text, only those templates whose key is a hypernym of, or equal to, the
synset at the current position in the parsed component list (allowing for skipping of
fixed token patterns such as ``where'') are examined.  This is done by constructing all
hypernyms for the current token (if it is a possible verb or noun) and using each as a
key.

All positions in the template list must match, for a match to be declared.  If the text's
parsed component list is longer than the template's list, all subsequence matches are
attempted (only one has to succeed for a match to be declared).  If the template item is a
fixed token (e.g. ``where''), that token must match exactly.  If the template item is a
verb synset $VS$, together with its tense, the set of synsets for the corresponding
position in the text must contain at least one synset which is a hyponym of $VS$, and the
tenses must match.  Finally, if the template item is a set of noun synsets $SNS$, then for
a match to occur, at least one member of that set must be a hypernym of (or equal to) one
of the synsets in the set of synsets for the corresponding position in the text.
Allowance is also made for skippable items: parsed components whose POS set contains one
or more of [\textit{adjective, adverb, unknown}]. For example, if a token's parsed
component POS set contains {\em adjective} and {\em noun}, then that token can be skipped
in the search for a match (if it itself does not take part in the match), and the next
noun set examined for a match.  In {\em ``He likes blue cheese''}, the token {\em blue},
taken alone, could be a noun or an adjective, so the matching process will both attempt to
make it part of the match, or will skip it if that match fails.

\subsubsection{The TAL Training Process}
\label{sec:talTraining}
In Train phase, the POS (noun or verb or neither) is confirmed with the Teacher, and the
Teacher is then asked to identify the single best matching synset for each noun or verb
thus identified.  As described above, to lighten the cognitive load on the Teacher, TAL
``guesses'' the correct synset to use to confirm with the Teacher, based on previous
Teacher inputs.  Also as mentioned, the Teacher confirms any compound phrases found by TAL
(if the Teacher declines the confirmation, the phrase is left in non-compound form), and
is asked to identify any compound phrases TAL missed; any such phrases are then added to
the taxonomy, with the Teacher's guidance.

Training can be done either in batch mode (which requires a set of positive and negative
labeled examples), or in free form mode (where the Teacher simply enters example phrases
instantiating the predicate they wish to teach).  We summarize below the iterative steps
involved in training a TAL model, in batch training mode:

\begin{enumerate}
\item A small number of labeled positive and negative phrases are first input for
  training.
\item TAL's first task is to elicit from the Teacher the predicate they wish to train.
  First, all predicates that match one or more of the positive training examples are
  listed, and the Teacher is asked to select one or ``Train another predicate''. If the
  latter is chosen, TAL lists all the predicates it knows about, along with an option to
  train a new predicate. If no existing predicates are found, the Teacher must train a new
  predicate.
\item TAL presents the next positive example to the Teacher, lets them know if the
  text already matches the predicate (i.e. matches a template containing the predicate),
  and asks them if they wish to train on this example. If there are no more positive
  examples, training is complete.
\item If Yes, TAL asks them to enter a phrase inspired by the input text, that captures
  the predicate being trained.  This allows the Teacher to zero in on just that part of
  the text that should fire the predicate, and to enter related text that should also
  fire the predicate, if desired.
\item TAL then shows any compound nouns and verbs it found, and asks for confirmation.
  Any that are confirmed as incorrect are dissolved back into single token form.
\item TAL asks if it missed any compound nouns, and adds them as indicated by the
  Teacher.  For the rest of the discussion, we will refer to both individual tokens and
  compound phrases as ``tokens''.
\item TAL asks the Teacher for the meaning of any tokens it doesn't know, and
  adds the new definitions to its WordNet ontology.  Thus the ontology is grown as needed,
  and these additions are shared between Teachers.
\item TAL assigns a set of possible parts of speech (POSs) to each token.  It only
  keeps tokens that are a question adverb, or that could possibly be a noun or verb.
\item If the POS set is ambiguous (still contains more than one POS), TAL asks the
  Teacher if the token is a noun, verb, or neither.
\item TAL lists the synsets for each noun or verb and asks the Teacher to pick one.  It
  then lists the possible generalizations (hypernyms) of that synset and asks the Teacher
  to generalize by picking one, if appropriate.
\item Once a unique set of synsets or strings has been identified, TAL constructs the
  corresponding template from them, and adds the predicate to that template.
\item TAL shows the template to the Teacher and asks them if it looks correct.  If they
  say no, TAL returns to step [4].
\item TAL then runs the new template over the negatively labeled training samples.  If
  this results in any false positives, TAL shows them to the Teacher and asks if they
  still want to keep the template. If so, the loop continues from [3] above; else, it
  continues from [4], but also gives the Teacher the opportunity to move to the next
  training sample (i.e. go to step [3]).
\item After the iterative training process is complete, the Teacher can read and modify
  the learned templates file (a text file in YAML format).
\end {enumerate}

We found the last step - editing the templates file - to be a simple and very useful
exercise: note that this can only be done with such a transparent (correctable and
interpretable) system.  We also allowed the Teacher to edit the templates on the fly, as
described in the next Section.

Thus the teaching process can result in new synsets (or nodes) being added to, or deleted
from, the taxonomy.  Examples of both of these are given in Section \ref{sec:FullTraining}
below.

\subsubsection{Editing the TAL Model}
After training, TAL's transparency allows us to manually edit the model.  This process can
be very efficient, especially when the number of templates is small, compared to adding
labeled data and retraining an SML model.  We followed the following procedure: 

\begin{enumerate}
\item We extended each template's sets of noun synsets, based on noun synsets occuring in
  the other templates.  Thus, for example, if one template allowed ``get.v.01'' followed
  by ``currency.n.01'', and another allowed the same verb synset but was missing a
  following ``currency.n.01'', we added it\footnote{This could all be done during the
    initial training process, but the instantiation used did allow parallel templates to
    develop (templates containing the same verb but different following or preceding noun
    synsets).}.
\item We removed any resulting duplicate templates.  
\item Over-generalization is checked.  For example, for the intent model trained below, we
  noticed during the teaching process that TAL suggested ``get.v.01'' as a more general
  version of ``buy.v.01''.  We found that replacing the latter by the former did indeed
  improve performance on the training set, for some templates, but for others, many false
  positives were generated.
\item We manually checked, using the online version of WordNet \cite{Fellbaum:1998}, that
  the synsets chosen previously made sense, again removing any resulting duplicates.
\end{enumerate}

\subsubsection{Testing TAL}
Test phase for TAL is simpler and involves no Teacher interaction.  The text is
normalized, compound phrases are identified using the (possibly extended) WordNet, and all
the templates are matched against the text, using the template matching logic described in
section \ref{sec:Templates}. If a template fires, TAL asserts that all of its associated
predicates are true. In the experiments described below, we had only one predicate; a test
example was declared positive if the predicate was true for that example.

\subsection{Experiments: Learning Curves}
\subsubsection{Data}
We generated training and test data from Cortana logs as follows: we searched for user
inputs containing one of more of the strings shown in Table \ref{tab:stargateQueries}.

\begin{table}[!ht]
\centering
\begin{tabular}{cccc}
\toprule
`` buy `` & `` buying `` & `` bought `` & \\
`` get `` & `` got `` & `` purchas'' & `` acquir'' \\
\bottomrule
\end{tabular}
\caption{Strings used to filter user inputs from the Cortana logs.}
\label{tab:stargateQueries}
\end{table}

Note the lack of a space after tokens `` purchas'' and `` acquir'', resulting in matches
for user inputs containing these strings as stems.  In addition, we adder filters to
restrict the user inputs to either speech or SMS texts; to inputs issued on Windows
devices, and to inputs issued to Cortana (as opposed to other Windows services).  Running
these queries over 18 days of logs from May, 2016 resulted in 15,310 records.  Uniquing
this data resulted in 12,690 records.  We also took the most recent 200 (distinct) queries
in the logs as unbiased background data.  We shuffled the 12,690 records and then took the
first 800.  We then labeled the resulting combined 1000 sentences as indicating the user's
intent to buy, or not.  The remaining 11,890 records were kept for use with the LUIS
active learning experiments.  Because some of the sample sets were very small (e.g. the
process only resulted in 14 queries containing `` acquir''), we split each individual
dataset in two (e.g. the `` acquir'' positives, and `` acquir'' negatives, were each split
in two) to create train and test sets.  This resulted in the totals shown in Table
\ref{tab:numSamples}\footnote{The sizes in Table \ref{tab:numSamples} add up to 999
  because one non-English query had slipped though the filters, which we later decided to
  drop since the language of the problem domain is English.}.

\begin{table}[!ht]
\centering
\begin{tabular}{ccc}
\toprule
	& \multicolumn{2}{c}{Sample Sizes} \\
\cmidrule(r){2-3}\vspace{-3mm}\\
	& Train & Test \\
\midrule
Positives & 78 & 82\\
Negatives & 417 & 422 \\
\bottomrule
\end{tabular}
\caption{Train and Test Sample Sizes}
\label{tab:numSamples}
\end{table}

\subsubsection{TAL and LUIS: Learning Curve Experiments} \label{sec:LearningCurve}
As mentioned above, we used the LUIS system \cite{WilKamMokMilZwe:2015} as a state of the
art machine learning baseline.  LUIS uses a logistic regression model with bag of words
features (inverse document frequencies) and the number of tokens as inputs.  LUIS also
gives the ability to use domain-specific dictionaries; we experimented with this but did
not find improved performance on this task.  Internally, LUIS performs 10 fold cross
validation to choose the optimal model parameters, and then trains on all the training
data, using the found parameter values.

Since constructing the learning curves required training 490 models for LUIS, in order to
make the comparisons both feasible and as fair as possible, for these experiments we used
both systems in maximally automated mode.  For LUIS this simply meant that no active
learning was used, while for TAL, we used a simplified version of the full system: the
user was not allowed to delete items from the ontology; they were not given the option of
editing the learned templates; and the templates used were a simplified version where each
synset set was constrained to be a singleton.  For TAL, full training (on 495 training
samples; see section \ref{sec:FullTraining} below) resulted in a total of 99 templates
(compare this with 24 templates, for the full system below trained on the same data, when
sets of noun synsets were allowed; and 21 templates if verbs are further combined into
sets where possible).

\begin{figure*}
\centering
\subfloat{\includegraphics[width=5 in]{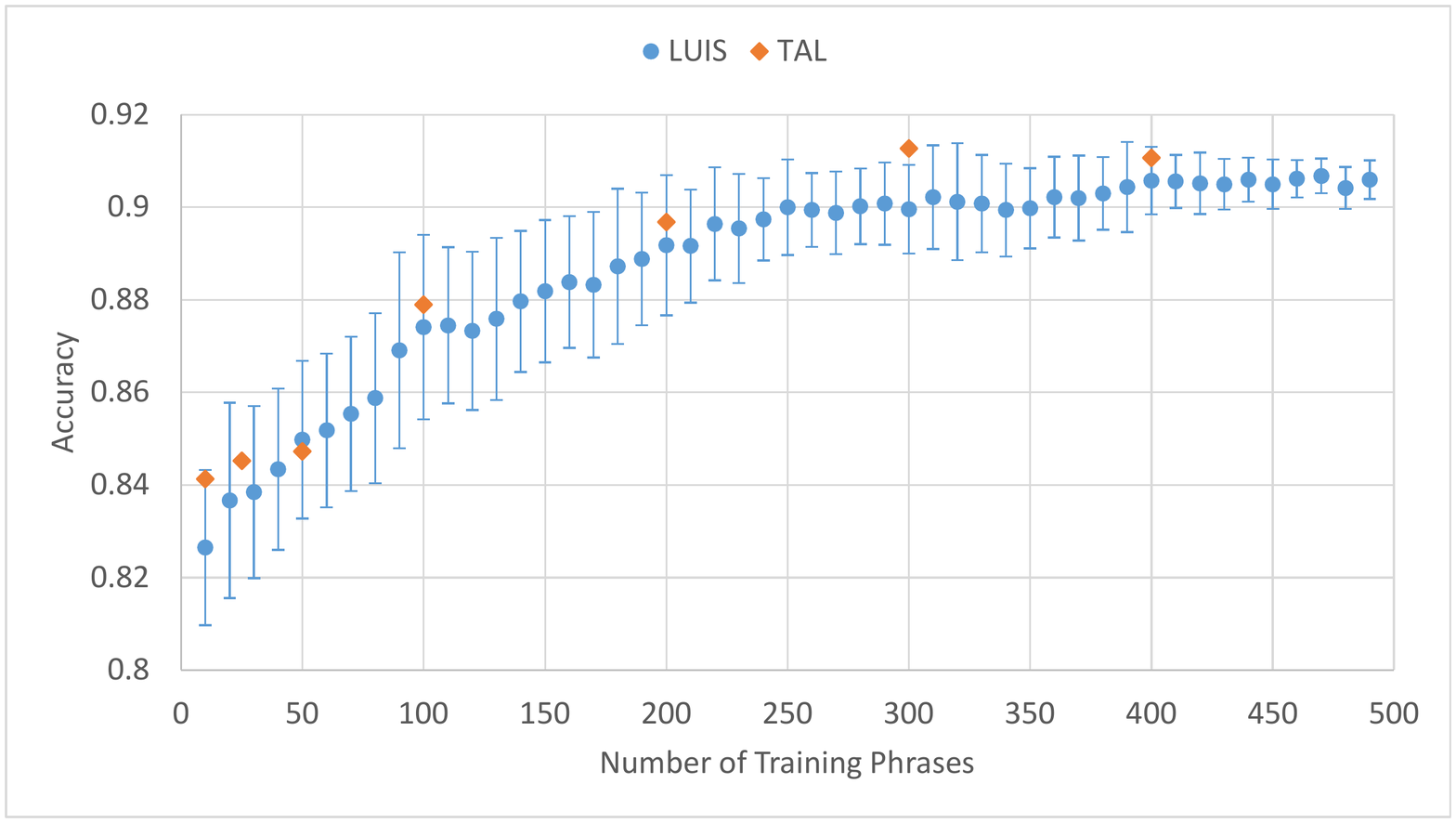}}
\\
\vspace*{-1.5 in}
\subfloat{\includegraphics[width=5 in]{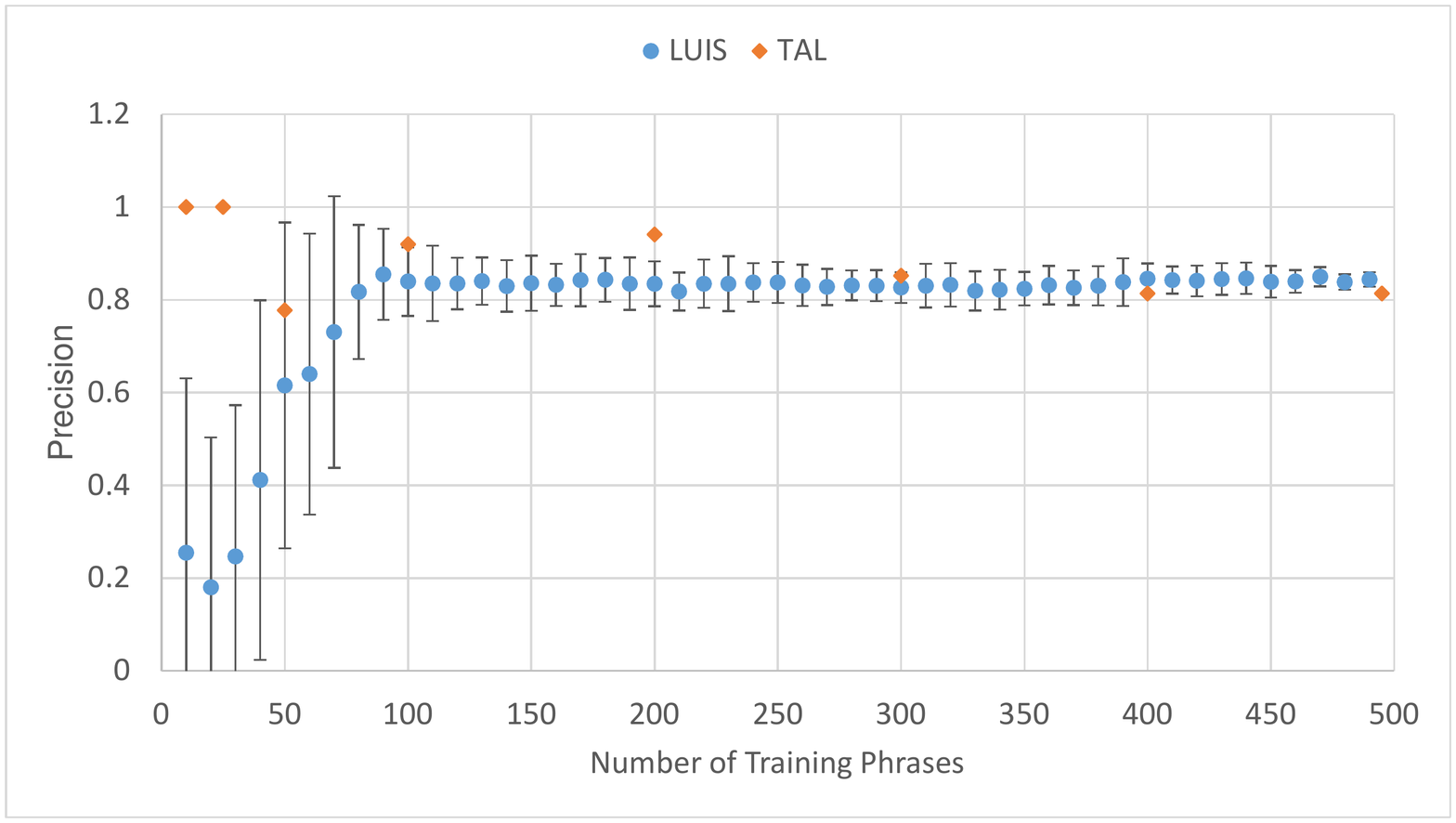}}
\\
\vspace*{-1.5 in}
\subfloat{\includegraphics[width=5 in]{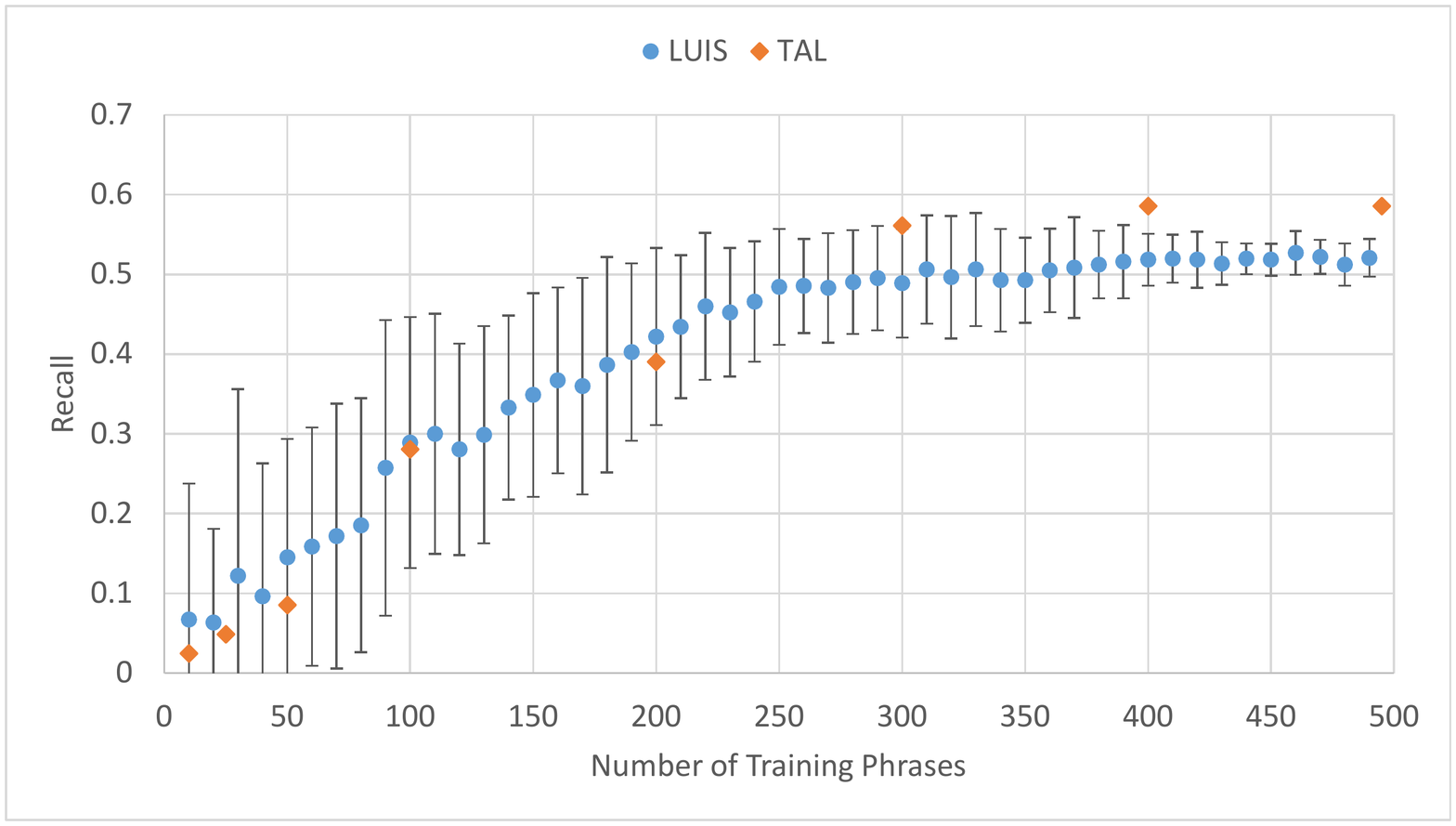}}
\caption{Accuracy, Precision and Recall Training Curves for LUIS and TAL.}
\label{fig:trainingCurves}
\end{figure*}

Figure \ref{fig:trainingCurves} shows the training curves for accuracy, precision and
recall on the 504 sample test set as a function of training set size.  For the LUIS
experiments, training set sizes were increased by 10 samples from one experiment to the
next, starting at 10 samples.  For TAL, the data was split into sets of size 20, 40, 60
and 78 positives.

\FloatBarrier
\FloatBarrier

\subsection{Experiments: Full Training} \label{sec:FullTraining}
We used the same training data as above.

\subsubsection{LUIS Baseline Results}
LUIS supports Active Learning (AL) via its 'Suggest' option. The user can upload a set of
unlabeled samples and then enter Suggest mode. LUIS will present up to 10 of the unlabeled
samples that it decides will most benefit the model to verify, along with its current
prediction for them. The user changes the prediction if necessary, or accepts the current
label. After the set of samples is labeled, LUIS retrains its model, then presents the
next set of up to 10 unlabeled samples.

After training LUIS initially from the full 495-sample test set, we ran it on the full
test set, then performed active learning using the 11,890 unlabled samples.  We did three
runs of active learning, creating 25 newly labeled samples, and ran it on the test set
again. Table \ref{tab:luisWithAL} shows these results along with TAL's scores on the test
set.

Note that the LUIS team made significant improvements to its intent prediction classifier
after we had performand the learning-curve experiments above.  This resulted in a pre-AL
baseline with significantly higher recall and somewhat less precision than before. Adding
active learning to LUIS increased its precision with no change in recall.

\subsubsection{TAL Results}
After training on the first 20 sentences, the manual editing process described above
resulted in reducing the number of templates from ten to seven.  This template collapse
pattern repeated, until after training on all the data, we arrived at 24 templates.  All
the templates found for the fully trained TAL user intent model are shown below, in Table
\ref{tab:classifier}.  In addition, since the model is fully transparent, we could also
check the other user-entered data, as shown in \ref{tab:wordNetOverlay}.

We also noticed that the 24 templates could easily be further reduced to 21 if verb
synsets are combined into sets where possible.  See Section \ref{sec:classifier} and Table
\ref{tab:classifier} for details.

\subsubsection{Comparison}
Compared to LUIS with active learning, TAL is significantly higher in precision and lower
in recall, for a marginal increase in overall accuracy.  A natural next step would be to
tune LUIS for the same precision and then measure its recall, and compare again.  However,
these results alone demonstrate that comparable performance can be achieved with a fully
transparant, succinct, and editable model.  Because of these advantages, we believe that
TAL could take advantage of more training data to increase accuracy arbitrarily (up to the
level of label noise).

\begin{table}[h!]
\centering
\begin{tabular}{llll}
\toprule
Experiment & Precision & Recall & Accuracy \\
\midrule
LUIS (no AL) & 0.800 & 0.683 & 0.919 \\
LUIS (with AL) & 0.824 & 0.683 & 0.925 \\
TAL & 0.911 & 0.622 & 0.929 \\
\bottomrule
\end{tabular}
\caption{Comparison of TAL's precision and recall to LUIS with and without active learning.}
\label{tab:luisWithAL}
\end{table}

\subsubsection{The Complete TAL Intent Detection Classifier}
\label{sec:classifier}
We present here the entire classifier that TAL learned to detect {\em intent to buy}.
First, the Teacher removed a single synset from the WordNet taxonomy.  TAL uses a YAML
file to track such removals, and in this case the file contains the single mapping {\em
  get: get.v.22}, meaning that the 22nd verb synset listed for the token {\em get} is to
be removed from the graph.  This was done to remove a loop in WordNet that causes TAL to
generalize incorrectly\footnote{get.v.22 (purchase) is a hyponym of buy.v.1 (purchase,
  acquire) which is a hyponym of get.v.1 (acquire); the occurrence of ``get'' being a
  hyponym of ``get'' prevented the Teacher from using ``buy'' as a special meaning of
  ``get''.}.

Table \ref{tab:wordNetOverlay} shows the Teacher-taught extensions to WordNet, with tokens
on the left and their mappings on the right.  The table distinguishes {\em concepts} and
{\em instances}: all mappings not denoted by the string {\em InstanceOf} are concepts.
Table \ref{tab:nounSet} shows a noun set referred to by many of the templates, referred to
as $SNS$ (for {\em shared noun set}).  Finally Table \ref{tab:classifier} shows the 24
templates used.  There, for example, {\em cost.v.01} denotes the first verb synset listed
in WordNet for the token {\em cost}; {\em PPF} is shorthand for past, present or future
tense, and {\em PF} for present or future tense.  Recall that the pronoun ``I'' is mapped
to the special synset tal\_narrator\_i.n.01, and ``you'' is mapped to the special synset
tal\_audience\_you.n.01 (see Section \ref{sec:templates}).  Finally sets of noun synsets
are denoted by the curly braces.  All templates fire the same ``intent to buy'' predicate,
which is therefore not shown.

It is a striking fact that the entire classifier can be written with complete transparency
(with consequent full interpretability and correctability) in one page.

\begin{table}[!ht]
\centering
\begin{tabular}{cc}
\multicolumn{2}{c}{ } \\
\cmidrule(r){1-1}
\cmidrule(l){2-2}
app & application.n.04 \\
at\&t & company.n.01 \\
at\&t hotspot & (InstanceOf) wireless local area network.n.01 \\
best buy & shop.n.01 \\
black ops 3 & (InstanceOf) computer game.n.01 \\
blu ray & videodisk.n.01 \\
cheeseburgers & food.n.01 \\
ge portable water dispenser & dispenser.n.01 \\
hoverboard & plaything.n.01 \\
killer instinct & (InstanceOf) computer game.n.01 \\
opal automobile & car.n.01 \\
sd cards & circuit board.n.01 \\
taco bell & restaurant.n.01 \\
\cmidrule(r){1-1}
\cmidrule(l){2-2}
\end{tabular}
\caption{The Teacher-taught extensions to WordNet.}
\label{tab:wordNetOverlay}
\end{table}

\begin{table}[!ht]
\centering
\begin{tabular}{cc}
\toprule
artifact.n.01 \\
commercial document.n.01 \\
currency.n.01 \\
dish.n.02 \\
domestic animal.n.01 \\
food.n.01 \\
food.n.02 \\
game.n.03 \\
license.n.01 \\
music.n.01 \\
plant.n.02 \\
precious metal.n.01 \\
software.n.01 \\
vehicle.n.01 \\
wireless local area network.n.01 \\
\bottomrule
\end{tabular}
\caption{The set of nouns shared by multiple templates, denoted by {\em SNS} in
Table \ref{tab:classifier}.}
\label{tab:nounSet}
\end{table}

\begin{table}[!ht]
\begin{adjustwidth}{-2cm}{}
\centering
\begin{tabular}{ccccc}
\toprule
cost.v.01 (PPF) & buy.v.01 (PF) & \{$SNS$\} \\
get.v.01 (PF) & \{dish.n.02\} \\
``how'' & buy.v.01 (PF) & \{$SNS$\} \\
``how'' & \{tal\_audience\_you.n.01\} & buy.v.01 (PF) & \{$SNS$\} \\
\{tal\_narrator\_i.n.01\} & achieve.v.01 (PF) & get.v.01 (PF) & \{$SNS$\} \\
\{tal\_narrator\_i.n.01\} &  buy.v.01 (PF) & \{$SNS$\} \\
\{tal\_narrator\_i.n.01\} &  desire.v.01 (PF) & buy.v.01 (PF) \\
\{tal\_narrator\_i.n.01\} &  desire.v.01 (PF) & get.v.01 (PF) & \{$SNS$\} \\
\{tal\_narrator\_i.n.01\} &  expect.v.01 (PF) & get.v.01 (PF) & \{$SNS$\} \\
\{tal\_narrator\_i.n.01\} &  want.v.01 (PF) & get.v.01 (PF) & \{$SNS$\} \\
\{tal\_narrator\_i.n.01\} &  refer.v.02 (PF) & get.v.01 (PF) & \{$SNS$\} \\
\{place.n.02\} & \{district.n.01\} & buy.v.01 (PF) & \{$SNS$\} \\
\{place.n.02\} & \{district.n.01\} & \{district.n.01\} & buy.v.01 (PF) & \{$SNS$\} \\
\{place.n.02\} & get.v.01 (PF) & \{$SNS$\} \\
remind.v.01 (PF) & \{person.n.01\} & get.v.01 (PF) & \{$SNS$\} \\
\{tal\_audience\_you.n.01\} & commend.v.04 (PF) & buy.v.01 (PF) & \{$SNS$\} \\
``what'' & \{$SNS$\} & get.v.01 (PF) \\
``what'' & \{$SNS$\} & \{tal\_narrator\_i.n.01\} & get.v.01 (PF) \\
``what'' & \{prerequisite.n.01\} & buy.v.01 (PF) & \{$SNS$\} \\
``where'' & buy.v.01 (PF) & \{game.n.01\} & \{artifact.n.01\} \\ 
``where'' & get.v.01 (PF) & \{$SNS$\} \\
``where'' & get.v.01 (PF) & cook.v.03 (PF) & \{food.n.02\} \\ 
``where'' & \{tal\_narrator\_i.n.01\} & buy.v.01 (PF) \\
``where'' & \{tal\_narrator\_i.n.01\}, \{tal\_audience\_you.n.01\} & get.v.01 (PF) & \{$SNS$\} \\
\bottomrule
\end{tabular}
\caption{The templates used by the classifier. Each row, left to right, forms a single
  template.  Note that most are quite general; some are specific to the training data
(e.g. the template with two \{district.n.01\}'s arose from the sentence {\em Cheapest
    place in Saint Louis Missouri to buy X}); and some lie in between (e.g. the
  template with the \{game.n.01\}, \{artifact.n.01\} synsets arose from {\em Where to buy
    golf shoes?} but would also apply to e.g. {\em Where to buy a tennis racquet?}}
\label{tab:classifier}
\end{adjustwidth}
\end{table}
\FloatBarrier

%% file: fdl.tex
\newpage
\section{Factored Dialog Learning}
\label{sec:FDL}
The language skills required of an APA can be broadly classified as the generation and
comprehension of language\footnote{By ``the APA comprehends X'' we simply mean that the
  phrase X triggered the desired actions by the APA, and did not trigger any undesired
  actions.}.  To model this, we designed two corresponding dialog structures which we call
\emph{Ask-How-To-Say} and \emph{Train-Predicate}.

\subsection{The Ask-How-To-Say Dialog Structure}
In the FDL framework, the language teaching stage occurs after the Designer has completed
their design for the class of applications and created the corresponding module
specification (this will be covered in the next section).  In the language teaching stage
the Teacher teaches the system the language it will need to converse with the User (beyond
any language that can be shared across all apps in the class; that too is specified by the
Designer).  Throughout the rest of this section we will use {\em Event Reminders} as an
example class of applications.

\emph{Ask-How-To-Say} is a dialog between the system and the Teacher that is intended to
develop the system's language generation skills.  The system asks the Teacher explicitly
how to say something pertaining to a given, general scenario, and it then uses the
Teacher's response later, when interacting with the User.  However, the system is not limited
to simply parroting the Teacher's phrases.  For example, for the Event Reminders class,
the system extracts a verb phrase from the Teacher that it can use throughout its dialog
with the User, as in:

\begin{quote}
  \textbf{S\footnote{We use \textbf{S}, \textbf{T}, \textbf{U} to represent the utterances
      of system, Teacher, and User respectively.}:} \textit{I need to know how to refer to
    the type of reminders created by your app. If I tell the user, "I understand that
    you'd like me to help remind you when to X", what would be a good phrase for X? Please
    be sure that:}
	\begin{enumerate}
    \item \textit{X starts with a verb (because it is an action, and will be inserted as
                a verb phrase into many sentences to request input and confirmation from
                the user).}
	\item \textit{X refers to the type of reminder rather than a specific instance of the
                reminder; for example, "feed the fish", but not "feed the fish once a
                day".}
	\end{enumerate}
\end{quote}

Thus when the Teacher is creating an Events Reminder app for tracking medications, for
example, they might type:

\begin{quote}
  \textbf{T:} \textit{take your medications}
\end{quote}

This verb phrase would then be used by the system in multiple different interactions with
the User.

Another way that the Designer can reduce the cognitive load on the Teacher is by making
judicious use of defaults.  Continuing with the meds tracking example, it might go like
this:

\begin{quote}
  \textbf{S:} \textit{I will ask the user to provide some optional notes for their
    reminder by saying ``What would you like to name this particular reminder?''. Does
    that sound good?}\\
  \textbf{T:} \textit{Yes.}
\end{quote}

\noindent Then later, when in dialog with the User, 

\begin{quote}
  \textbf{S:} \textit{What would you like to name this particular reminder?}\\
  \textbf{U:} \textit{Take aspirin.}
\end{quote}

\noindent This dialog would then trigger the APA to add an event to the User's calendar labeled
``Take aspirin''.

The Module Specification Language is described below, but we just make the connection to
the MSL here by pointing out that in the MSL YAML file, the above actually appears as
follows:

\FloatBarrier
\begin{figure*}[!ht]
\begin{mdframed}
  \texttt{ParamValueTurn}:\\
  \phantom{xx} \texttt{Confirm:} I will ask the user to provide a name for their new
  reminder by saying ``$\langle \mathrm{ask\_event\_name} \rangle $''. $\langle
    \mathrm{DoesThatSoundGood} \rangle$\\
  \phantom{xx} \texttt{Loopback:} $\langle \mathrm{OKEnterSentence} \rangle $\\
  \phantom{xx} \texttt{Param:} $\langle \mathrm{ask\_event\_name} \rangle $\\
  \phantom{xx} \texttt{Type:} STRING\\
  \phantom{xx} \texttt{Default:} What would you like to name this particular reminder?\\
\end{mdframed}
  \caption{Example \texttt{ParamValueTurn} that suggests to the Teacher default language to use to ask the User
    for a name for their reminder app, and accepts and stores an alternative form if the
    Teacher so desires.}
  \label{fig:param-value-turn-1}
\end{figure*}
\FloatBarrier

\noindent where the system asks the Teacher whether the default language is acceptable,
and if it is not, then asks them for the sentence that they would like to use instead.
This sentence is then stored by the system in the variable <ask\_event\_name>.  Note that
the system can also use its own variables (like <DoesThatSoundGood>, which in this case
translates to the string {\em ``Does that sound good?''}.  The MSL has several such
constructs that are intended to make the Designer's task easier, and they are described
below.

In User mode, the system gathers the data it needs for the running application by
asking the User questions, using application-specific language.  To keep the load on the
User as light as possible, the system at first prompts the User to simply enter the data
in free form, and then confirms its understanding with the User, and queries the User for
any missing data.  Thus, for example, for the Prescriptions Reminder app, the User can
just directly enter their prescription, and as long as all the required data is there, TAL
can then just confirm, and then populate the User's calendar.

In order to be able to do this, TAL needs to be able to parse, and to model, fundamental
concepts such as space, time, and number.  To this end we use \textit{Parsed Predicates},
as opposed to \textit{Learned Predicates} (see above). Parsed predicates are patterns that
are coded by the developer to parse text and extract key quantities. To make this
concrete, we briefly describe the F\# structures we use for the built-in parsed
predicates.  For example, TAL's code contains a {\em Frequency} module that collects
several active patterns; one is called {\em OncePerSecond} and models frequencies written
in the form ``Once per second'', ``Twice per day'', ``4 times an hour'', etc. (We adopt
the convention that the active pattern's name be like one of the patterns it matches, to
make the code easier to read).  We call these collections of active patterns, each
designed to detect a particular pattern in language, {\em microgrammars}.  At run time,
all microgrammars are scanned across the text, in a particular order: it is thus important
to keep the active patterns independent or, if one contains another, to call the more
specific version first.  For example, in our current code the active pattern {\em Hourly},
which matches ``hourly'', ``daily'', ``annually'' etc., is called after the active pattern
{\em OnceHourly}, which detects ``once hourly'', ``twice daily'', etc.  However, such
dependencies are rare: active patterns can be similar yet still independent.  For example,
the active pattern for {\em EverySecond}, which matches ``Every day'', ``In the morning'',
etc., is similar to the more specific pattern {\em EveryNtoMSeconds}, which matches
``every 3-4 seconds'', ``every 6 to 9 minutes'', etc., but these patterns are independent
(i.e. they will never both be triggered by the same text) and so can be used in the scan
in either order.

The parsed predicates scan the microgrammar active patterns over the text to fill their
slots.  For example, our built-in parsed predicate for {\em Frequency} has two slots, {\em
Period} (to model the base frequency, like ``weekly'') and {\em NumberOfEventsPerPeriod}
(to model the ``twice'' in ``twice daily'').

Thus TAL's built-in parsed predicates form a rudimentary world model.  The hard-wiring
raises the concern that this would limit scalability.  However, the concepts modeled are
fundamental, and can be shared across all apps: we argue that it makes sense to allow
limited hardwiring, to take advantage of the extreme succinctness of basic physical models
(imagine learning to count, from positive and negative examples only, with no underlying
model of the number line).  We touch on this issue again in Section
\ref{subsec:extensibility}.

\subsection{The Train-Predicate Dialog Structure}
While Factored Dialog Learning is centered around factoring the language out of the
problem using built-in parsed predicates, we still need Teacher Assisted Learning to learn
any predicates the app needs that are not covered by the built-ins. In our running example
of the Events Reminders, the only place learned predicates are used is to detect which app
the User wants to run.  In general, apps will need more learned predicates: we minimized
the dependence on learned predicates here to simplify our exploration (and exposition) of
FDL.  So here, we just use TAL to learn when the User wants to run their app.  This will
be important in practice, for two reasons: the primary interface is expected to be speech
only; and the number of apps is likely to become large\footnote{AppBrain estimates that
the number of Android apps is currently at the 2.5 million
mark\cite{NumberOfAndroidApps:2016}.}, so listing them all is not practical (especially
with a speech-only interface).

For example, in the module specification for the event reminder class of apps, the
Designer can introduce a ``Train Predicate'' dialog:

\begin{quote}
  \textbf{S:} \textit{When the user wants to run an app, they will type a phrase that
    describes what they want to do, and I will find apps that match that intent. So, you
    need to teach me how to recognize when the user wants to run your app, by entering
    example phrases that the user might type to do this; for example, "feed the fish" to
    create a fish-feeding reminder.}\\
\newline
    \textit{I will try to generalize your examples, so that you don't have to type in too
      many.}\\ \textit{I will use what you tell me to build a "predicate", which describes
      the meaning I will look for in the user's text.}\\ 

	\textit{For each sentence, I will retrieve and display all existing templates that
      match for your sentence at one or more Verb positions. This will allow you to merge
      synsets from your new template into an existing template, remove synsets from
      existing templates, or remove existing templates.}
\end{quote}

\noindent Then, the Teacher can type in phrases like

\begin{quote}
  \textbf{T1:} \textit{track meds}
  
  \textbf{T2:} \textit{take my medications}
\end{quote}

Note that it is advantageous for the Teacher to zero in on key phrases, since they can be
detected in longer sentences which may also contain irrelevant details.  The process is as
described above in \fullref{sec:TAL}: the Teacher will specify the meaning of each
noun and verb according to the ontology, and, for example, the system will generalize
``medication'' to ``drug'', if the Teacher chooses to.  The system then will recognize
User input such as

\begin{quote}
  \textbf{U:} \textit{tell me when to take my pills}
\end{quote}

\noindent since there exists a synset for ``pill'' which is a hyponym of ``medication'', and since
possessive pronouns are ignored by the template matching process.

\subsection{Overview of the Module Specification Language}
\label{sec:MSLspec}
The MSL (Module Specification Language) is specified in detail in \nameref{app:MSLspec}.
Here, we give an overview of the main ideas behind the MSL.

In order to extend the APA to support a new application domain, the Designer must create a
module specification script that supports the new class of applications\footnote{We use
the term ``module'' here to denote the class of applications being considered.}.

\subsubsection{Script Blocks}
A module specification script consists of four sections, or blocks, listed below. We
will describe the notion of a \emph{turn} in more detail below, but at a high level, it
is a series of system-User interactions that together aim to acquire a single piece of
information from the User, or to deliver a single piece of information to the User.

\begin{description}
  \item[\texttt{Name:}] A string that defines the name of the module (for example, {\em
    Event Reminders}).
  \item[\texttt{Initialize}] block: A set of app-independent constants used to streamline
    the interactions with the Teacher (and, less commonly, the User).
  \item[\texttt{Teach}] block: Both the \texttt{Teach} and \texttt{Use} blocks consist of
    a series of two kinds of turn: predicate detecting turns, and parameter value turns.
    At a high level, the \texttt{Teach} block is a turn-taking structure, where the
    Designer designs a series of questions intended to leverage the Teacher's language
    skills to teach the application the language comprehension and generation skills that
    are to be used in User mode.
  \item[\texttt{Use}] block: This is a turn-taking structure that obtains and parses free
    User input in order to create the actual entity. In the Events Reminder example, a
    predicate detecting turn is used to parse free User input describing their reminders,
    and a series of parameter value turns confirms the values found or to ask the User to
    fill in missing values (such as the length of time the reminders should run for).
\end{description}

\subsubsection{The ``Turn'' in detail}

The {\em Turn} is the key atomic building block in a module specification.  Depending on
what type of information is to be gathered or delivered, one of four core turn types are
used: \texttt{Prompt}, \texttt{ParamValueTurn}, \texttt{TrainPredicateTurn}, and
\texttt{TestPredicateTurn}.

\begin{description}
  \item[\texttt{Prompt:}] A message that the system will show the Teacher/User.
  \item[\texttt{ParamValueTurn:}] The system can ask the Teacher/User a question and
    obtain the answer and/or confirm with the Teacher/User that some value is correct.
    Any parameter value thus obtained can be referenced in the subsequent turns (or in the
    current turn if it is used after the value is assigned).  Figure
    \ref{fig:param-value-turn-2} shows an example that asks for and confirms the start
    date for a reminder app.
  \item[\texttt{TrainPredicateTurn:}] Used to train the templates needed to form a learned
    predicate detector, where the Teacher supplies example phrases to train the desired
    \texttt{Predicate}. The Teacher types an example phrase that should fire the
    predicate, and the system engages in dialog with the Teacher to select the correct
    parse, synsets, and their generalizations for the identified terms or compound terms.
    The system uses this information to create a template which it associates with the
    predicate.
  \item[\texttt{TestPredicateTurn:}] The system displays a \texttt{Question} to prompt
    the User to provide needed information: for example, for a medications tracker app in
    the Event Reminder class, the User might be prompted to input their prescription.  The
    system them parses the User's input to detect matches with both parsed and learned
    predicates.  The components of any matching predicates (such as frequency, time, and
    duration) are mapped to the parameters defined in the \texttt{PredicateParamMappings}
    field, which can then be referenced in subsequent turns. Values that are present are
    used to populate a confirmation question ({\em ``You would like your reminder at 7:00 a.m.
    Is that correct (Y/N)?''}); missing values require the user to enter that value
    ({\em ``Please tell me what time you would like your reminder''}) before confirming it.
\end{description}

\begin{figure*}[!ht]
\begin{mdframed}
  \texttt{ParamValueTurn:}\\
  \phantom{xx} \texttt{Question:} $\langle \mathrm{ask\_start\_date\_language} \rangle$\\
  \phantom{xx} \texttt{Confirm:} You would like to start on $\langle \mathrm{start\_date} \rangle$. Is that correct?\\
  \phantom{xx} \texttt{MistypeFollowup:} What date is that? You could enter ``today'', ``tomorrow'', or a specific date such as ``July 19, 2016''.\\
  \phantom{xx} \texttt{Param:} $\mathrm{start\_date}$\\
  \phantom{xx} \texttt{Type:} DATE
\end{mdframed}
  \caption{Example \texttt{ParamValueTurn} that asks for and confirms the start date of a reminder app.}
  \label{fig:param-value-turn-2}
\end{figure*}

\subsubsection{Other Turn Types in the MSL}
In addition to the above four core \textbf{Turn} types, the MSL provides composite
\textbf{Turn} types that define iteration and conditional execution, auxiliary
\textbf{Turn} types that support these, and \textbf{Turn} types that explicitly control
parameter values. We briefly describe these here; see \nameref{app:MSLspec} for complete
details.

\begin{description}
\item[\texttt{SetParamValues:}] Allows the Designer to set parameter values directly
  in the script. 
\item[\texttt{RemoveParamValues:}]  For deleting previously assigned parameter values.
\item[\texttt{ScriptConditionalActions:}] Equivalent to an if-else-then dialog flow,
  based on the boolean value of a condition written in a simple conditional grammar that
  allows testing whether a script parameter is set or has an infinite value, comparing numeric
  values (=, >, >=, etc.), and parentheses for precedence.
\item[\texttt{UserConditionalActions:}] Like \texttt{ScriptConditionalActions}, but the condition
  is the boolean result of a question presented to the User. For example, in the Reminder
  application, if all necessary slots are filled, the User will be presented with a single
  sentence to confirm; if {\em No} is returned, then the script presents the series of
  ParamValueTurns.
\item[\texttt{ParamValueOrConstantTurn:}] Like \texttt{ParamValueTurn}, but allows the
  Teacher to enter a constant value if so doing is more appropriate for the app they are
  creating.  For example, a Teacher creating a birthday reminder app would not want to
  have the app ask the User how often their reminder should fire.
\item[\texttt{SetConstantValues:}] Sets script variables to constant values (such as
  those set by ParamValueOrConstantTurn) if they were not already set (such as by a
  TestPredicateTurn). 
\item[\texttt{NIterations:}] Equivalent to a for-loop.
\item[\texttt{EndIterationAction:}] For NIterations; the equivalent of a break in a for-loop,
  with a parameter to specify ending either the innermost or all nested loops (if any).
\item[\texttt{EndScriptAction:}] Ends the script, with a parameter indicating success or
  failure.
\item[\texttt{NoAction:}] A no-operation; functions as an empty branch of an ``if''.
\end{description}

\subsection{Case Study: Adapting the Event Reminder Module}

To evaluate the generalizability of the FDL framework, we performed a small (3
participants) user study to determine whether ordinary English speakers can easily and 
effectively function as Teachers, creating reminder applications based in the Event
Reminder application class (module). One Teacher participant and one new participant
then operated as Users, testing whether these applications could be used to easily and
intuitively create reminders for their domains.

\subsubsection{Teacher Task Description}

Using the Event Reminder Module, we assigned the participants to create two reminder
applications: a yearly reminder (such as a birthday) and a regular reminder (such as going
to the gym). We provided an informational document that described both their role as a
Teacher (creating an application that a User will use to create actual reminders) and the
general form of the prompts they would receive and the answers they should supply. Other
information was contained in the prompts themselves, many of which showed examples taken
from a prototype Medication Tracker application.

Subject A had some linguistic background, and created a birthday reminder application with
an early version of the FDL system; feedback from this session was incorporated into a new
version. With this second version, Subject A re-did the birthday reminder application and
then created a gym reminder application. Subject B, who had no linguistic background,
did the gym reminder application first, then the birthday reminder application.
Neither subject had any programming or application-development experience.

Subject C is a software developer with no significant linguistic background, and used a
third version of FDL that had been modified based upon the feedback from Subjects A and
B, creating first a regular ``Call your Friend' reminder application, and then a Wedding
Anniversary reminder application. Additionally, for Subject C, the ``medication tracker'
examples in the prompts were largely replaced with a ``feed the fish'' imaginary
application, and some additional features were added to the FDL system (these are
discussed below in the relevant subsections).

We observed the teachers as they performed the tasks and asked them to provide feedback
when complete. In particular, we asked whether they felt the system was expressive enough
to create the desired application, whether the task was harder or easier than expected,
and whether any parts of the task were particularly difficult, confusing, or easy.

\subsubsection{Teacher Task Results}

The distinction between the Teacher creating an application vs. a User who will use that
application to create the actual instances (in this case, reminders) is a key focus of
FDL. Subject A's initial use of the system incorrectly started from the User perspective,
for example by using an actual reminder description ("my father's birthday") instead of a
reminder creation action (e.g. "remember a birthday") for the application activity
("you would like me to help you remember to ..."). The documentation and prompts were
modified to clarify this distinction and this was not an issue for Subject A's subsequent
pass. Subject B still exhibited some confusion in this, although much less than Subject A
did initially. Subject C had only minor confusion about this. A detailed tutorial
combined with experience using the app will be sufficient to clarify this important
distinction.

Subject A provided additional valuable feedback to streamline and clarify the system
between the first two iterations, in particular in generalizing from the medication
reminder examples to a different reminder topic, clarifying how multiple reminders
(delimited by "then") are detected, clarifying when a response should be "yes/no" vs. a
sentence or phrase, improving the flow of specific and generalized synset selection
(including eliminating duplicate selection requests), and merging or omitting some steps.
After this iteration of user feedback and system revisions, the first-use experience for
Subjects B was much more straightforward. Building upon feedback from Subjects A and B,
additional features (discussed below) were added to address specific pain-points, which
led to an improved first-use experience for Subject C.

Subjects A and B found the gym reminder much easier to understand and create, for the
following reasons:
\begin{itemize}
\item The gym reminder shares its schedule structure with the example medication
reminder; both are geared towards daily, while the birthday reminder is annual.
\item The gym reminder, like the medication reminder, refers to a specific, concrete
event; the birthday reminder is more ambiguous, as it may be for sending an email or card,
buying a gift, planning or attending a party, taking a trip to visit, or simply an
indefinite reminder to "remember the birthday."
\item Some questions have only one applicable value in the reminder domain, such as
{\em "how often do you want the reminder sent?"} for the birthday reminder. Both subjects
suggested that these should allow the Teacher to set a constant value such as "annually"
and not present the question to the User.
\end{itemize}

The FDL system allows the Teacher to provide multiple example sentences for the same
Learned Predicate. For Subjects A and B, FDL required that the Teacher explicitly create a
new predicate for the application, and create a different one when the meaning changed.
Neither subject understood this clearly from the prompts. Subject B, lacking a linguistic
background, found the concept of predicates confusing until more explanation was provided,
and also did not initially realize that the example sentences were to be supplied as the
User would enter them. Additionally, Subject B felt some of the instructions (such as
requiring that the phrase referring to the application start with a verb) were
unnecessarily restrictive.

This feedback led us to make the following changes before Subject C's session:
\begin{itemize}
	\item The ability to specify constant values was added. Subject C happily used this
to specify that reminders for the Wedding Anniversary would occur once annually, forever.
	\item Because each application should define a single function or a small set of closely
related functions, the creation and selection of Learned Predicates was removed in favor
of a single predicate per application, reducing complexity for the Teacher.
	\item Additional explanation was added in the prompts. For example, to address Subject
B's feeling that requiring a Verb to start the phrase referring to the application was too
restrictive, additional text was added to explain that this is due to how it will be used
in subsequent prompts.
\end{itemize}

These changes resulted in much less confusion and a more streamlined experience for
Subject C. However, subject C was still puzzled in a couple of areas, such as not knowing
the definition of a compound, and in seeing the name of a script variable. These were 
subsequently modified for clarity. Subject C also mentioned that some of the prompts
were long enough to be difficult to read. Moving much of the material from the prompts
to a tutorial that includes both a detailed walk-through of Teaching an application and
a parallel demonstration of that application in action during the Use phase would 
improve both the first-use and, by reducing clutter, subsequent uses by the Teacher.

Both subjects felt that the turn-by-turn dialog was helpful in building up the app, and
were much more comfortable with the process at the conclusion of their tasks. However,
for the Teacher, a GUI rather than the command-line interface would reduce clutter
(such as when selecting synsets and generalizations, which can take multiple screens)
and could add support for Undo functionality, which would reduce the need for
multiple confirmations.

\subsubsection{Teacher Task Summary}

While there were some initial mistakes by all subjects, the FDL system's evolution led
to a much better experience as testing proceeded. All subjects became more comfortable
as they developed experience using the system. With a better tutorial providing a more
detailed walk-through of both Teacher and Use phases and a GUI that reduces screen
clutter and supports Undo, we believe that Teachers who have solid English skills and
no programming or application-development experience will be able to quickly learn how
to use the FDL system to create applications, especially if they have some linguistic
background.

\subsubsection{User Task Description}

Subject A and a new Subject D (who has linguistic and machine-learning experience) used
the Medication Reminder script to create actual reminders to take medication.

\subsubsection{User Task Results}

Subject A again went first, and created reminders for each of 10 prescriptions. Having
operated twice previously as a Teacher, Subject A was by now familiar with the goals
of the process. There was some confusion when being asked to specify a single value
for a range (for example, ``every 4 to 6 hours'' must be converted to a single ``every
N hours'' within that range). Here the dialog was not clear as to the context, and there
was no clear error message when it retried on out-of-range values. There were also some
missing recognitions of abbreviations for intervals. Once these were encountered and
understood, the process proceeded quickly, and the last few reminders were completed
easily.

Subject D used the system after Subject A's feedback was incorporated. Subject D entered
only a single prescription and found the process quite straightforward; the only negative
feedback was that the number of confirmations slowed the process down.

\subsubsection{User Task Summary}

Both subjects felt the turn-taking process made it easy to provide the necessary
information to create the reminder. Both subjects felt that the confirmations could be
streamlined to make the process flow more smoothly.

%% file: discussion.tex
\section{Discussion}
In this section, we look back, and assess how things went; in the next, we look ahead, and
map out some possibe routes upward, from our newly established base camp, for developing
fully transparent and scalable approaches to AI.

Our main result is that, with the tools we've developed, it's straightforward to build a
user-intent detector that is fully transparent, editable, and succinct, and that performs
as well as a state of the art machine learning approach.  Furthermore, the tools
themselves do not rely on sophisticated NLP methods such as semantic role labeling or
consituency or dependency parsing; they do not employ statistical methods at all - yet.
Thus not only are the results transparent, so are the tools used to achieve them.  Finally
the ideas should apply equally well to detecting other meanings in text.

However we found that our initial hope, that the Teacher would only need a clear
understanding of the app they wish to build, and of their own native language, was only
partially fulfilled; while it's certainly possible to so restrict the teacher's role, we
found that allowing them to edit the text files that contain the templates, and the
extensions and deletions in the taxonomy, was very useful.  This requires an extra
level of training for the Teachers.  But if the hope of just a few designers, feeding
hundreds of teachers, who can then support millions of users, comes to pass, perhaps it's
not too much to ask a little more of the Teacher role.

A more serious issue is that we made little headway towards our ``Every teacher benefits
from what any teacher teaches'' North Star.  The built-in parsed predicates are shared,
and certainly, teachers can reuse predicate detectors that others have built.  But true
reusability will likely have to employ hierarchical structures of predicates.  We will
discuss some ideas for this in the next section.

An earlier version of TAL used gazetteers to identify US cities and retailers.  We found
that we did not need these gazeteers, at least for the ``intent to buy'' detection task.
Later versions may need them.  However, compound phrase handling does need to be improved.
We found that allowing both paths (i.e. searching for a match by treating a possibly
compound phrase as compound, and also as non-compound) introduced training errors, as did
treating possibly compound phrases as definitely compound.  It seems that to solve this
robustly we will need to use context; if the phrase itself could be a compound noun, check
that a noun phrase is a possible part of speech, given the surrounding tokens; similarly
for verbs; and only employ the 'try both paths' trick if the compoundness is truly
ambiguous.  Note that this still does not require a full NLP parse of the text.

Finally the ideas outlined in this paper need to be further checked against more tasks and
on other datasets.  Our hope is that the benefits resulting from a fully transparent
approach will warrant such explorations.

%% file: ideas.tex
\section{Some Ideas for Future Work}
\label{sec:ideas}
For the discussion in this section it is helpful to name two phases of learning that TAL
can do: ``wake'' phase, in which TAL learns directly from a human, and ``dream'' phase, in
which no human is involved.  In dream phase, for example, TAL could learn either by
leveraging large, unlabeled datasets, or by directly making more sense of what it has
already learned.

\subsection{Wake Phase Error Correction}
We will describe ideas for error correction using large unlabeled datasets below, but we
can also ask the teacher for confirmation of consequences of their choices that thay may
not have anticipated.  If the teacher has arrived at a template using labeled data, TAL
could run it in real time over a large unlabeled dataset, show the resulting matches to
the teacher, and then allow them to modify the template accordingly.  False positives are
thus easy to control.  False negatives pose a harder challenge. For these, perhaps the
simplest approach would be to run the unlabeled data through both TAL and a statistical
system such as LUIS, then have the teacher manually inspect any samples on which the two
systems disagree and update the templates as needed.

\subsection{Dream Phase Error Correction}

\subsubsection{Template Error Correction Using Unlabeled Data}
Suppose that a teacher has taught TAL an overly general template intended to fire the
predicate ``person ingests food''.  Suppose that their template is (schematically) as
illustrated in Table \ref{tab:synsetsAndRoles}, where the synsets of the template are in
the first column and the corresponding roles defined by the teacher (i.e. the
corresponding predicate's components) are in the right column. This template
overgeneralizes because, for example, it will fire for the phrase ``building eats
person''.

\begin{table}[ht!]
\centering
\begin{tabular}{lr}
\toprule
Synset & Role \\
\midrule
entity.n.01 & ingester\\
eat.v.01 & act of ingesting \\
entity.n.01 & ingested\\
\bottomrule
\end{tabular}
\caption{Synsets and Roles}
\label{tab:synsetsAndRoles}
\end{table}

One way to leverage a large, offline, unlabeled dataset to check that a template does not
overgeneralize is as follows.  First, run the template over the dataset, making a record
of all matches.  For each match, increment a counter in the node in the WordNet-based
taxonomy, where the counter is labeled by its component name in the corresponding
predicate (e.g. ``ingester'').  (Note that a given token will correspond to multiple nodes
since at test time, we don't know its correct synset.)  The taxonomy hypernym tree (for
each component of the predicate) can thus be represented as a heat map.  If the template
is too general, then the heat map will contain cold spots at, or close to, the leaves.
(The token ``building'' will rarely occur as an ``ingester'' in the dataset).  This
computation gives us a distribution over the nodes of a subtree of the taxonomy for each
predicate component, where the root of the subtree is that component's synset.  We can
extend this reasoning to handle templates with sets of synsets, but for clarity here we
consider components that correspond to a single synset in the template being tested.

For concreteness consider just the first component, and call its tree $T$.  If some
subtree $S\subset T$ has nodes that are sufficiently ``cold'', one could try to find a
subtree $S'\subset T$ that maximizes the number of ``hot'' nodes and minimizes the number
of ``cold'' nodes.  If necessary, one could split the original template into two or more,
such that each has only ``hot'' nodes, and all ``hot'' nodes in $T$ are covered (this fits
naturally into our template model that uses sets of noun synsets).  This process is
necessarily statistical, since rare phrases do occur (buildings can eat people, in fiction
or metaphors).

An interesting line for future work is to take the brakes off this process and simply ask:
given a large, unlabeled text dataset, how can one choose $N$ templates for which the
amount of matching text is maximized, but such that amount of text for which pairs of
templates match is minimized?  If $N=1$, presumably a template could be chosen that will
cover most, or all, of the data.  For $N\ge 2$, templates compete to explain the data.  In
this way, one might approach the fully offline learning of templates.  How far can one get
by using only the taxonomy and large unlabeled datasets, in this way?  Could some
predicates arise naturally from the resulting templates?  Could such a set of templates
help inform us on how to build a suitable hierarchy of predicates?

\subsubsection{Template Unification}
Similarly a large unlabeled dataset can be used to identify templates that are essentially
the same, even if their sets of synsets differ, by the similarity of their heat maps.
First, two templates that are candidates for unification could be found by automatically
comparing pairs of heat maps.  Those templates might then be unified if a template could
be found that maximally agrees with the combined heat maps of the individual templates.  A
process like this will likely be needed to identify similar templates proposed by
different teachers, if the automated checking process used during training (``your input
phrase fired this template: combine or add?'') fails.  Similar ideas can be applied to
identify when the predicate being trained is likely the same as one already learned.  Note
that these ideas are already instantiated, to some extent, in TAL: any phrase input by a
Teacher is tested against all existing predicates, and the Teacher is alerted if there is
a match; similarly, TAL currently allows the unification of parts of templates, if those
templates share a verb synset.

\subsection{Extensibility}
\label{subsec:extensibility}

\subsubsection{Multiple Languages}
TAL's language dependence currently resides in the WordNet taxonomy; a ``microgrammars''
file that instantiates the built-in parsed predicates; an ``English POS'' file that lists
parts of speech that only occur via small numbers of tokens (e.g. pronouns, and question
adverbs); the function that maps text to lemmas and tense, using the Slang Map; the
function that maps annotated lemmas to the 'parsed components list'; and two supporting
files, of verb declinations and nouns with their plurals.  Clearly the adherence to
transparency comes with a development cost when it comes to adding new languages.  WordNet
is supported for languages other than English \cite{WordNetForOtherLanguages:2016}, and
one can certainly envision using other, similar taxonomies.  Extending templates and
predicates from one language to another requires more than just having a mapping of
synsets available, since different languages can express the same concept in very
different ways.  Even so, it may be possible to use machine translation systems to
accomplish this automatically.  If one has a large database of paired phrases for the two
languages, then the templates in the new language could be learned automatically, and its
predicates' component names could be translated similarly.  This idea, combined with a
given mapping of synsets from one language to the other, could also be used to learn the
mapping of the predicate components to their corresponding template components in the new
language.

\subsubsection{TAL's World Model}
We think of the built-in parsed predicates - e.g. detecting properties of time, space, and
number - as forming the basis of TAL's world model.  Currently the structure containing
the predicates is flat, but it's clear that a hiearchy will be needed: velocity needs the
concepts of space and time, acceleration needs velocity and time, etc.  It seems likely
that a hierarchy would also be required to make learned predicates scalable.  If one
entity is about to eat the other, then the two entities must be colocated.  If two people
are married then they must know each other, must have made a joint commitment, etc.

One way to encode logical relationships is to have a directed graph whose nodes are
negatable predicates (by a negatable predicate, we just mean a predicate with an added
negation flag: thus {\em he is not eating fish} would fire the {\em Ingests(person, food)}
predicate with its {\bf Not} flag set).  If a node is true only if its children are all
true, then we have the logical connectives {\bf And} and {\bf Not}, which form a
functionally complete set (a set from which any truth table can be constructed).  It may
however be more convenient to explicitly model the {\bf Or} connective by instead having a
directed bigraph in which nodes are either predicates or connectives, and for which every
node has an attached Boolean variable.  For example, if several predicate nodes are
connected to an {\bf Or} node, the latter is true only if one or more of those predicates
is true.  Such a graph is one way of implementing logical inference, and such a hierarchy
would also help with the interpretability of the overall model.  One can apply similar
ideas to those above, to detect logical relations between predicates automatically. Thus
if a set of templates and predicates is run over a very large dataset, and predicate $A$
is always found to be true whenever predicate $B$ holds, then the relation $B \Rightarrow
A$ could be added to the graph.  It is more likely, however, that a probabilistic way of
modeling logic, such as Markov logic networks \cite{RicDom:2006}, would be required.

For scalability, if a new parsed predicate must be added, and it can be defined in terms
of existing parsed predicates (for example, speed or velocity in terms of position and
time), ideally one would define the new predicate in terms of the others in a script file,
without having to write new code.

\subsubsection{Extending to Large Numbers of Templates and Predicates}
TAL already uses a simple mechanism to limit its search: roughly speaking, only those
templates that are indexed by the token currently being examined are tested for a match
for the following tokens.  The search can be further limited by partitioning templates
using context.  For example, when TAL is awaiting user input to determine which app to
run, it knows it's in a particular state, and so it can safely use templates that would
overgeneralize in other situations: so for example a template containing the single verb
``exercise'' can be used to select a ``go to the gym'' reminder app, and not used
elsewhere.  More generally, TAL can represent its inner state using a finite state
automaton, and have only subsets of templates (and their predicates) available for each
state.  This could be extended to its interactions with the user: TAL could model the
user's state using a Markov chain, where each state again links to a limited set of
templates, and only those templates corresponding to states whose probabiliy exceeds a
threshold would be tested for matches.

\subsubsection{Extending the MSL}
We currently define a number of \texttt{Turn} types in MSL (see
\nameref{app:MSLspec}). These are intended to be generic and flexible enough to
accommodate all the necessary question-answering and control-flow operations needed by a
Designer. Supporting domain-specific \texttt{Turn} types would require a plug-in model for
FDL to understand which actions to take for which questions.

We may consider supporting module reuse through some form of \textit{module inheritance},
allowing a module to extend a \textit{base module}. The \textit{base module} might
include all the required metadata-collecting operations such as those in the
\texttt{Initialize} block, so the the \textit{derived modules} do not need to repeat them.
This is similar to the superclass constructor execution when constructing subclass
instances in object-oriented programming. Similarly, we might extend the script blocks to
allow a base module to define a set of of \texttt{Turns} that could be ``called'' by a
derived module.

\subsection{Other Benefits of Model Transparency}
Suppose you have several statistical models that were trained (possibly using different
data) for the same task.  How best to combine them?  Model averaging often works well.
But that's a hack and it requires running all the models.  With a fully transparent model
such as TAL, the models (i.e. their templates) can be directly combined, resulting in a
model with the benefits of all the trained models but that is both more compact and
efficient than using all of the models independently.  This is another way that
transparency can contribute to scalability.  Multiple Teachers can do their thing on their
own data, training the same predicate (task), and the results can be combined.  Being able
to handle the inputs of many teachers is a key requirement for scalability.

\subsection{The Machine Comprehension of Language}
In \cite{Burges:2013}), the following practical definition of the machine comprehension of
text was proposed:

\begin{quote}
A machine comprehends a passage of text if, for any question regarding that text that can
be answered correctly by a majority of native speakers, that machine can provide a string
which those speakers would agree both answers that question, and does not contain
information irrelevant to that question.
\end{quote}

We can recast this in terms of predicates as follows:

\begin{quote}
A machine comprehends a passage of text if, for every (possibly implicit) assertion made by
the text, as identified by a majority of native speakers, that machine asserts the
corresponding predicate to be true, and it does not assert any other predicates to be
true.\footnote{Note that this definition also covers negations.}
\end{quote}

Note that some 'false negatives' are easy to detect automatically: if no predicate fires
for a passage, and if we know that the passage has some meaning that should have been
detected (as would be the case, for example, in parsing a passage for general question
answering), then the machine necessarily does not comprehend that text.

The above definition gives us a way to measure how well a system comprehends a dataset of
text samples.  We could first crowd source the labeling, asking workers to write down all
assertions (predicates) made by each sample text, and then determine the overlap between
the human generated predicates, and those predicates declared to be true by the system
when run over the text.  Errors introduced during labeling could be controlled by using
multiple workers for each sample, and by asking a further set of workers to verify any
outlier claims.  We might further test the system's world model by distinguishing those
predicates that follow directly from the text, and those that are implied indirectly by
it.

This measurement strategy would of course work for any system that makes assertions via
predicates, giant black-box neural nets included.  However, as we have argued above, we
believe that the fully transparent, interpretable, correctable, predicate-based approach
presented in this paper at least provides us with a good starting point - a base camp -
for further investigations into scaling AI.

%% file: ack.tex
\section{Acknowledgements}
We thank Jason Williams, Paul Bennett and Richard Hughes for many valuable discussions and
for their support of this work.  We also thank Vishal Thakkar, the Stargate Support team,
and Hisami Suzuki for their help and guidance in our generating the Cortana dataset.

\noindent
This work is the culmination of several years of explorations. We thank Erin Renshaw for
her steadfast help with the earlier work.  We thank John Platt for his vision in
supporting big-bet, long-term research at MSR, which freed us to (repeatedly) fail.  We
additionally thank Eric Horvitz, Jeannette Wing and Harry Shum for their leadership and
vision in supporting long term research in general and this work in particular.

%% file: appendix_1.tex
\newpage
\section{Appendix 1: Specification of the Module Specification Language}
\label{app:MSLspec}

This section provides the detailed specification of the Module Specification Language
(MSL) used by the Designer to create module scripts for FDL. A brief overview was given
above in \fullref{sec:MSLspec}.

\subsection{YAML Basics}

YAML\footnote{\url{http://yaml.org}} is a serialized data representation format that can
be edited easily by humans and parsed by a program. Our module specification language uses
Yaml as a representation medium. In this section, we discuss the elements of YAML that
are used in the example Medication Reminder Module script.

\textit{Scalar} is the basic data type, which can represent a single value, e.g. a number
or a string. Type inference is automatically done most of the time in Yaml; Table
\ref{tab:sampleYamlTypes} contains some examples.

\begin{table}[h!]
\centering
\begin{tabular}{cc}
\toprule
Value & Type \\
\midrule
25' & integer \\
``25'' & string \\
25.0' & float \\
Yes & boolean \\
\bottomrule
\end{tabular}
\caption{Examples of YAML type inference.}
\label{tab:sampleYamlTypes}
\end{table}

\textit{Mapping} and \textit{sequence} are two basic ways to composite scalars.
\textit{Mapping} is similar to the \textit{map} or \textit{dictionary} data structures
in most programming languages; values are referenced by unique keys, and the key/value
pairs are unordered. The key and the value are separated by \textit{a colon and a space},
for example
\begin{verbatim}
    name: Tom
    age: 15
    married: Yes
\end{verbatim}
is a mapping. In Yaml, the indentation really matters; all entries in a mapping must have
the same indentation.

\textit{Sequence} is similar to the \textit{list} data structure in most programming
languages, which means the values are ordered and referenced by the index (position in the
list). Each entry begins with \textit{a hyphen and a space}. For example
\begin{verbatim}
    - Tom
    - Jerry
    - Jack
\end{verbatim}
is a list of three entries. Again, indentation matters; the hyphens must have the name
indentation.

An empty sequence or map is indicated by:
\begin{verbatim}
    MyList: []
    MyMap: {}
\end{verbatim}

\vbox{
Complex objects can be represented by a combination of mappings and sequences. For
example, a sequence of instances of a data structure containing the fields ``name'',
``age'', and ``married'' would be:
\begin{verbatim}
    - name: Tom
      age: 15
      married: Yes
    - name: Jerry
      age: 14
      married: No
    - name: Jack
      age: 12
      married: No
\end{verbatim}
Again, note that the hyphens are indented at the same level, and each hyphen is the
start of an instance of the structure.}

By default, YAML concatenates all lines together in a single value. You can use a special
sequence starting with `|' to preserve linebreaks and whitespace; `|-' suppresses the
final one linebreak.
\begin{verbatim}
        Question: This becomes
                one line, with one space between "becomes one".
        Question: |-
                This remains three lines with
                1. Item One
                2. Item Two
\end{verbatim}

The comment indicator in Yaml is `\#'; anything from this to the end of the line is
ignored.

\subsection{Script blocks}

A module specification script partitions the specification into four sections, or blocks,
listed below. We will describe the notion of a \emph{turn} in more detail below, but at a
high level, it is a series of system-User interactions that together aim to acquire a
single piece of information from the User, or to deliver a single piece of information to
the User.

\begin{description}
  \item[\texttt{Name:}] A string that defines the name of the module (for example, {\em
    Event Reminders}).
  \item[\texttt{Initialize}] block: A set of app-independent constants used to streamline
    the interactions with the Teacher (and in principle with the User, although most
    variables used in User mode are set in the \texttt{Teach} block).  For example, our
    Event Reminders module defines the $\langle \mathrm{OKEnterSentence} \rangle$ string
    constant to have the value {\em ``OK, then please enter the sentence you would like me
      to use here.''} and it's used six times in the \texttt{Teach} block.  The
    \texttt{Initialize} block can also be used to ask the teacher any questions needed to
    acquire metadata about the application; this data is then saved using reserved keys.
  \item[\texttt{Teach}] block: Both the \texttt{Teach} and \texttt{Use} blocks consist of
    a series of two kinds of turn: predicate detecting turns, and parameter value turns.
    At a high level, the \texttt{Teach} block is a turn-taking structure, where the
    Designer designs a series of questions intended to leverage the Teacher's language
    skills to teach the application the language comprehension and generation skills that
    are to be used in User mode.  Both turn types can also be embedded in conditional
    blocks.  In the Events Reminder example, a predicate turn is used to elicit from the
    Teacher sentences that exemplify the language that the User is likely to input in order
    to identify this application from the pool of apps available to the User; this lets the APA
    know which app the User wants to run.  In the parameter value turns, the Teacher specifies,
    for example, the language used to elicit a name for the series of reminders that the User
    is creating.  The Designer can also elicit from the Teacher any language that is
    likely to be repeated (as in the example above); for example, the Designer might want
    to extract a general verb phrase from the Teacher, to refer to the type of reminders
    created by the app (for example, ``take your medications'' for a meds reminder app);
    this verb phrase is then referred to throughout the remainder of the script (in both
    Teach and Use blocks).
  \item[\texttt{Use}] block: This contains a set of sub-blocks named for the action being
    performed on the instance. In the Event Reminder module, the only action defined is
    \texttt{create}, when creating the reminder. This could be extended with, for example,
    \texttt{edit}, which reads an existing reminder and allows the user to edit it. Each
    \texttt{Use} sub-block is a turn-taking structure that obtains that parses free user
    input in order to execute the action on the entity; for the \texttt{create} action, this
    is to create the entity. In the Events Reminder example, a \textbf{TestPredicateTurn}
    parses free User input describing their reminders and obtains parsed predicates,
    and a series of parameter value turns confirms the values found or asks the User to
    fill in missing values (such as the length of time the reminders should run for).
    Thus, for an individual reminder, ideally a single predicate detecting turn receives a
    sentence from the User that contains all the information needed to create the
    reminder, and the system then confirms the information in a single turn. If the user
    chooses to change some of this information, or if some necessary information is
    missing, further parameter value turns are run to gather the needed data.
\end{description}

Note that the natural sequential nature of an MLS script means that one can create
variables early in the script that can be used throughout the rest of the script, in all
blocks.

\subsection{Script Parameters}

Values are obtained from the Teacher or User and stored in parameters; parameters also
contain the values to be presented to the user.

\subsubsection{Parameter Names and References}
\label{sec:ParameterNames}

The parameter should be named using only A-Z, a-z, 0-9, underscore ("\_"), or period
("."), and is case sensitive. There are some special name formats with the underscore
that should be reserved for a specific purpose; see \fullref{sec:SpecialParameterNames}.

In the following discussion and subsections, we occasionally use as an example a parameter
named \texttt{par}.

A parameter value can be referenced in the subsequent turns (or in the current turn if it
is used after the value is assigned) using the syntax \texttt{\$\{par\}}. The process of
converting from this name representation to its value is referred to as expansion.
Depending on the context of the parameter reference, the parameter value may be converted
to the various types (e.g. to string if used in screen printing).

Parameter names can be built up from other parameters into a composite name. For
example, \texttt{\$\{frequency.\$\{i\}.period\}} first obtains the value for \$\{i\}, for example
\texttt{0}, and then forms the name \$\{frequency.0.period\}, which can then be looked up
directly. The bracketing \$\{\} is required if expansion is to be done, and otherwise should
not be present. In particular, when a parameter's name is to be used rather than its
value, expansion should not be done for the entire name, but it may have to be done for
embedded names if the parameter is a composite. This is discussed in more detail below.

\subsubsection{Special Parameter Names}
\label{sec:SpecialParameterNames}

We suggest reserving names that begin with an underscore for parameters with special
meaning. In particular, names using all uppercase letters with two underscores at both the
beginning and end should indicate ``external'' values, such as the state of the script, 
parameters that have a special meaning to FDL, or a value that an application will look
for. For example, section \fullref{sec:FindingTheApp} describes how some special
parameters are used to find the application the user wants to run.

Table \ref{tab:SpecialParameterNames} shows the current set of special names in FDL.

\FloatBarrier
\begin{table}[h!]
\centering
\begin{mdframed}
\begin{tabularx}{\textwidth}{l|X}
  Name & Meaning \\
  \hline
  \hline
  \_\_NAME\_\_ & The name of the application. Another application may look for this. \\
  \hline
  \_\_IS\_EDITING\_\_ & Indicates that the script is being edited rather than being a
newly created script. This means that some parameter values may already be present, which
will let the script designer present a more relevant dialog. \\
  \hline
  \_\_ACTION\_DESCRIPTION\_\_.\textit{action\_name} & \texttt{\_\_ACTION\_DESCRIPTION\_\_}
is the base name for a collection of parameters of type \texttt{STRING}, one for each
action in the \texttt{Use} block, that is used by FDL to give the user a friendly
description of what this action does. For example,
\texttt{\_\_ACTION\_DESCRIPTION\_\_.create} provides a description of
the \texttt{create} action of the Use block, such as ``create a reminder to go to the
gym''. \\
  \hline
  \_\_ACTION\_CONFIRMATION\_\_.\textit{action\_name} & \texttt{\_\_ACTION\_CONFIRMATION\_\_}
is the base name for a collection of parameters of type \texttt{STRING}, one for each
action in the \texttt{Use} block, that is used by FDL to ask the user a question to
confirm that this script and action is really what the user intends to do. For example,
\texttt{\_\_ACTION\_CONFIRMATION\_\_.create} provides a confirmation question for the
\texttt{create} action of the Use block, such as ``do you want to create a reminder to go
to the gym?''. \\
  \hline
  \_\textit{anything} & Indicates a parameter intended to hold a value for ``internal''
use by a \textbf{SetParamValues Turn}, such as reducing duplicate strings, rather than
being obtained from the user. \\
\end{tabularx}
\end{mdframed}
\caption{Special Parameter Names in FDL.}
\label{tab:SpecialParameterNames}
\end{table}
\FloatBarrier

\subsubsection{Parameter Types}
\label{sec:ParameterTypes}

Table \ref{tab:ParameterTypes} lists the parameter types currently supported by FDL.

\FloatBarrier
\begin{table}[h!]
\centering
\begin{mdframed}
\begin{tabularx}{\textwidth}{l|l|X}
  Name & Type & Description \\
  \hline
  \hline
  STRING & StringValue & The user's answer is saved as it is without further language
understanding or information extraction. \\
  \hline
  NUMBER & NumberValue & Extracts a number expression from the user's
input and saves it as a float value. It supports digit expressions in English format
(e.g. "3.14") or English cardinals (e.g. "thirty five") for integers only. If the user's
input starts with "i don't know" or is "forever'', the \texttt{infinity} value is
assigned. NUMBERs can be compared to other NUMBERs or to INFINITY; see
\fullref{sec:ConditionalParams}. \\
  \hline
  none & NumberValueRange & A NUMBER range such as "2-3". This is created by a User
entering a string during a \texttt{TestPredicateTurn}; script turns require a single
value. \\
  \hline
  DATE & DateValue & Extracts a Date from the user's input. This supports a)
\href{https://msdn.microsoft.com/en-us/library/2h3syy57.aspx}{.NET DateTime.Parse}
compatible strings (e.g. "July 21, 2016"), b) special days (e.g. "today", "tomorrow",
"now"), c) days of the week (e.g. "Tuesday"), d) the day of the next week as in "next
Tuesday". \\
  \hline
  TIME & TimeValue & Extracts a Time from the user's input. This supports a)
\href{https://msdn.microsoft.com/en-us/library/2h3syy57.aspx}{.NET DateTime.Parse}
compatible strings (e.g. "8AM"), b) time expressed using English words (e.g. "three
thirty five"), c) offset by 12 hours if "pm", "p.m.", "evening", or "afternoon" is present. \\
  \hline
  INTERVAL & IntervalValue & Extracts a time interval from the user's input. The
supported units include "second", "minute", "hour", "day", "week", "month"
and "year" , and the value can be either digits (e.g. "3 years") or English words (e.g.
"three years"). We also support "half" (e.g. "one year and a half"). All interval
expressions are stored as a
\href{https://msdn.microsoft.com/en-us/library/system.timespan.aspx}{.NET TimeSpan}
object. The number of "months" and "years" is converted to "days" depending on the
current month and year, i.e. "one month" can be converted to "28 days" to "31 days". It
also supports an "i don't know" or "forever" answer from the user, in which case
\href{https://msdn.microsoft.com/en-us/library/system.timespan.maxvalue.aspx}{TimeSpan.MaxValue}
is used. INTERVALs can be compared to INFINITY; see \fullref{sec:ConditionalParams}. \\
  none & IntervalValueRange & An INTERVAL range such as "2-3 hours". This is created by a
User entering a string during a \texttt{TestPredicateTurn}; script turns require a single
value. \\
  \hline
  YESNO & YesNoValue & If you expect the user to answer with either "yes" or "no", then
you could use this parameter type. \\
  \hline
  NOMINAL & StringValue & If you have a list of all valid answer strings, you could
specify the parameter type `NOMINAL` and list the valid strings of the answer. \\
\end{tabularx}
\end{mdframed}
\caption{Parameter types.}
\label{tab:ParameterTypes}
\end{table}
\FloatBarrier

Here is an example of defining a parameter's name and type, for example in a
\textbf{SetParamValues} or \textbf{ParamValueTurn}:
\begin{verbatim}
Param: person_name
Type: STRING
\end{verbatim}

To define a parameter of \textbf{NOMINAL} type, specify the values as a mapping, which
should start on a new line with extra indentation, e.g.:
\begin{verbatim}
Param: fruit_category
Type:
  NOMINAL: [apple, orange, pear]
\end{verbatim}

\subsubsection{Conditional Operations on Parameters}
\label{sec:ConditionalParams}

FDL supports conditional comparisons to provide if-then-else logic for selecting which
\textbf{Turn}s to execute. Table \ref{tab:ConditionalParameters} lists the conditional
operations currently supported by FDL; these are the ones that were useful in developing
the Medication Reminder module, and more may be added in the future. These comparisons
are done in a \textbf{ScriptConditionalActions Turn}.

\FloatBarrier
\begin{table}[h!]
\centering
\begin{mdframed}
\begin{tabularx}{\textwidth}{l|l|X}
  Operation & Supported On & Description \\
  \hline
  \hline
  =, <>, >, <, >=, <= & NUMBER & Boolean result of the comparison. \\
  \hline
  IS [NOT] INFINITY & NUMBER, INTERVAL & Boolean result indicating whether the named
  parameter's value is or is not infinity. If the parameter is not set, an exception is
  thrown; check with IS SET first. \\
  \hline
  IS [NOT] SET & all & Boolean result indicating whether the named parameter has or has
  not been assigned a value. If the parameter has been removed by a RemoveParameterValues
  \textbf{Turn}, it IS NOT SET. \\
  \hline
  IS [NOT] NUMBER & all & Boolean result indicating whether the named parameter is or is
  not a NUMBER. \\
  \hline
  IS [NOT] INTERVAL & all & Boolean result indicating whether the named parameter is or is
  not an INTERVAL. \\
\end{tabularx}
\end{mdframed}
\caption{Conditional operations supported on parameters.}
\label{tab:ConditionalParameters}
\end{table}

Comparisons may be combined with AND and OR (which short-circuit), and parentheses may
be used for precedence.
\FloatBarrier

\subsubsection{Arithmetic Operations on Parameters}

FDL supports a limited set of binary operations on parameter values; the syntax is similar
to parameter references except that two parameter names (or a parameter name and a value)
are present in the expression and are separated by an operator. Currently, we support the
binary operators "+" and "-" between a `DATE` and an `INTERVAL` parameter, and
between a `NUMBER` parameter and a float value. This is illustrated in Table
\ref{tab:ArithOps}.

\FloatBarrier
\begin{table}[h!]
\centering
\begin{mdframed}
\begin{tabularx}{\textwidth}{l|l|l|X}
  Type 1 & Type 2 & Operator & Example \\
  \hline
  \hline
  DATE & INTERVAL & +,- & Assuming a \texttt{DATE} parameter \texttt{date} with value
\texttt{July 21, 2016} and an \texttt{INTERVAL} parameter \texttt{interval} with value
\texttt{3 days}, then \texttt{\$\{date+interval\}} results in the \texttt{DATE} value
\texttt{July 24, 2016}. \\
  \hline
  NUMBER & float & +,- & Assuming a \texttt{NUMBER} parameter \texttt{i} with value
\texttt{1}, then \texttt{\$\{i-1\}} results in the \texttt{float} value \texttt{0}. \\
\end{tabularx}
\end{mdframed}
\caption{Arithmetic operations supported on parameters.}
\label{tab:ArithOps}
\end{table}
\FloatBarrier

\subsection{Script Handlers}
\label{sec:Handlers}

A \texttt{Handler} combines a script name with an action name and is the means by which
the FDL system determines which scripts apply to a user's intended action. For a
discussion and example of how \texttt{Handler}s are used, see \fullref{sec:FindingTheApp}.
Table \ref{tab:HandlerProperties} describes the \texttt{Handler} properties.

\FloatBarrier
\begin{table}[h!]
\centering
\begin{mdframed}
\begin{tabularx}{\textwidth}{l|l|l|X}
  Property & Type & Required & Description \\
  \hline
  \hline
  ActionName & string & yes & The action to be performed when this predicate fires. \\
  \hline
  ScriptName & string & yes & The name of the script to execute \texttt{ActionName} in
  when this predicate fires. \\
\end{tabularx}
\end{mdframed}
\caption{Handler properties.}
\label{tab:HandlerProperties}
\end{table}
\FloatBarrier

\subsection{Turn Types}

A \textbf{Turn} in our context is a series of interactions that together aim to acquire
a single piece of information from the user or deliver a single piece of information to
the user. This is the atomic building block of an interactive application.

\subsubsection{Prompt}
\label{turn:Prompt}

This defines a message that the system will show the Teacher or User. It has no 
properties; the value of the Turn itself is the message to be displayed.

\noindent \textbf{Example:}
\begin{quote}
\begin{verbatim}
	Prompt: Let's edit your application.
\end{verbatim}
\end{quote}
\bigskip

\subsubsection{ParamValueTurn}
\label{turn:ParamValueTurn}

Presents a question to the Teacher or User and obtains the result value. There are 
numerous possible interaction flows, depending upon the presence or absence of the
referenced parameter and properties.

\FloatBarrier
\begin{table}[h!]
\centering
\begin{mdframed}
\begin{tabularx}{\textwidth}{l|l|l|X}
  Property & Type & Required & Description \\
  \hline
  \hline
  Description & string & no & Provides a description that is shown only once per execution
  of the step \\
  \hline
  Param & string & yes & The name of the parameter being asked for or confirmed. \\
  \hline
  Type & ParamType & yes & The type of the parameter, from \fullref{sec:ParameterTypes}. \\
  \hline
  Default & string & no & A default value to assign to the parameter if it is not set. \\
  \hline
  Question & string & no & A question to ask the Teacher or User to request the parameter
  value. \\
  \hline
  Confirm & string & no & A confirmation message to use when the parameter is set. \\
  \hline
  Loopback & string & no & If the user does not confirm when asked, then this is the prompt
  that is used to repeat the request for the parameter value. If it is not present, then the
  \texttt{Question} property is used. \\
  \hline
  MistypeFollowup & string & no & If the value entered by the user cannot be parsed into
  a value of \texttt{Type}, then this is a custom message to display; if it is not present,
  a general message is shown. \\
\end{tabularx}
\end{mdframed}
\caption{ParamValueTurn properties.}
\end{table}
\FloatBarrier

The FDL system will append ``(Y/N)?'' to the the \texttt{Confirm} question when it 
displays it to the Teacher or User.

\bigskip
\textbf{Examples:}
\bigskip

The simplest \textbf{ParamValueTurn} consists of only \texttt{Param}, \texttt{Type}, and
\texttt{Question} properties, where the system asks the user a question defined in the
\texttt{Question} property, converts the answer to the type specified in the \texttt{Type}
property, and assigns the result to the parameter named in the \texttt{Param} property.

\noindent \textbf{Example:}

\begin{minipage}{\linewidth}
\begin{quote}
\begin{verbatim}
	ParamValueTurn:
	  Question: Please name the script (e.g. "AMI Medication Tracker").
	  Param: __NAME__
	  Type: STRING
\end{verbatim}
\end{quote}
\end{minipage}
\bigskip

If there is a viable default value for the parameter, the script can confirm this value
with the Teacher or User without asking him or her the initial question. In this case,
the \textbf{ParamValueTurn} would have the \texttt{Default}, \texttt{Confirm}, and 
\texttt{Loopback} properties. The system \textit{first} assigns the value in the
\texttt{Default} property to the parameter named in the \texttt{Param} property, and then
confirms with the user using the question in the \texttt{Confirm} property. If the user
says "no" to the \texttt{Confirm} question, then the \texttt{Loopback} question will be
asked and the user will be asked to \texttt{Confirm} again. Usually the \texttt{Param}
should be referenced in the \texttt{Confirm}. 

\noindent \textbf{Example:}

\begin{minipage}{\linewidth}
\begin{quote}
\begin{verbatim}
	ParamValueTurn:
	  Default: 3 days
	  Loopback: Please tell me over what duration you'd like your reminders set?
	  Confirm: You would like to take medication for ${duration}. Is that correct?
	  Param: duration
	  Type: INTERVAL
\end{verbatim}
\end{quote}
\end{minipage}
\bigskip

A single \texttt{ParamValueTurn} can contain both a \texttt{Question} and a \texttt{Confirm}.
In this case, the system first asks the user the \texttt{Question} and saves the value in the
\texttt{Param} before \texttt{Confirm}ing with the user. If the user denies the confirmation,
then user will be asked the the \texttt{Loopback} question if it exists; otherwise the same
\texttt{Question} will be asked again. Either way, the \texttt{Confirm} is asked again and the
process continues.

If you specify a particular type, but the system couldn't recognize an instance of that type in
the user's input, then you could also define the \texttt{MistypeFollowup} property to give the
user more hints on what we expect the user to input.

\noindent \textbf{Example:}

\begin{minipage}{\linewidth}
\begin{quote}
\begin{verbatim}
	ParamValueTurn:
	  Question: When are you going to start taking the medication?
	  Confirm: You would like to start on ${start_date}. Is that correct?
	  MistypeFollowup: What date is that? You could enter "today", "tomorrow", or a specific date such as "July 19, 2016".
	  Param: start_date
	  Type: DATE
\end{verbatim}
\end{quote}
\end{minipage}
\bigskip

Another important feature of \texttt{ParamValueTurn} is that if \texttt{Param} has been
initialized in the previous turns, then the current turn will be skipped, \textit{unless
the existing value type doesn't match the required param type}, in which case the system
forces the type mapping with the user's help. For example, if we define a parameter
\texttt{duration} with the type \texttt{INTERVAL}, then the expected value type should
be \texttt{IntervalValue}; if we find the value of this parameter has the value type
\texttt{IntervalValueRange} (see \fullref{sec:ParameterTypes}),
then the system asks the user to select an \texttt{IntervalValue} from the
\texttt{IntervalValueRange} and assigns the updated value to the parameter
\texttt{duration}. A similar dialog occurs if a \texttt{NUMBER} parameter has a value
that is a \texttt{NumberValueRange}.

To summarize, the value is assigned to a parameter in the following order:

\begin{enumerate}
  \item[1]. Existing value assigned in the past, if any;
  \item[2]. The expanded \texttt{Default} property of the \textbf{ParamValueTurn};
  \item[3]. The User's input as response to the \texttt{Question}.
\end{enumerate}

\subsubsection{TrainPredicateTurn}
\label{turn:TrainPredicateTurn}

The \textbf{TrainPredicateTurn} leverages the Teacher's language skills to train the
templates needed to form a \texttt{LearnedPredicate} detector. When the system enters a
\textbf{TrainPredicateTurn} it asks the teacher to type an example phrase that should fire
the \texttt{LearnedPredicate} associated with the application being created.  The system
then engages in dialog with the Teacher to select the correct parse, synsets, and
their generalizations for the identified terms or compound terms, creates a template,
and associates the predicate with that template. The process of supplying example
sentences continues until the Teacher has no further phrases they wish to use; for
example, when every new phrase the Teacher enters is matched by an existing template.

The \textbf{TrainPredicateTurn} has the following properties:

\FloatBarrier
\begin{table}[h!]
\centering
\begin{mdframed}
\begin{tabularx}{\textwidth}{l|l|l|X}
  Property & Type & Required & Description \\
  \hline
  \hline
  PredicateName & string & yes & Names the \texttt{LearnedPredicate} associated with
  the application. The \texttt{PredicateName} property is usually named for the
  application, as there is currently only support for one \texttt{LearnedPredicate} in
  FDL. \\
  \hline
  Notes & string & no & Any additional free-form information associated with the
  \texttt{LearnedPredicate}. \\
  \hline
  \hline
  Handler & struct & yes & A structure containing the mapping between action names and
  the names of scripts that implement those actions. See \fullref{sec:Handlers} \\
  \hline
  ActionName & string & yes & The action to be performed when this predicate fires. \\
  \hline
  ScriptName & string & yes & The name of the script to execute \texttt{ActionName} in
  when this predicate fires. \\
  \hline
\end{tabularx}
\end{mdframed}
\caption{TrainPredicateTurn properties.}
\end{table}
\FloatBarrier

\noindent \textbf{Example:}

\begin{minipage}{\linewidth}
\begin{quote}
\begin{verbatim}
	- TrainPredicateTurn:
	    PredicateName: ${event_phrase}	# The phrase used to refer to the event
	    Notes: ""
	    Handler:
	        ActionName: "create"
	        ScriptName: ${__NAME__}		# The name of the current script
\end{verbatim}
\end{quote}
\end{minipage}
\bigskip

\subsubsection{TestPredicateTurn}
\label{turn:TestPredicateTurn}

The \texttt{TestPredicateTurn} is used in User mode to determine which predicates
match input from the user. The system displays the \texttt{Question} to prompt
the User to provide needed information: for example, for a medications tracker app in
the Event Reminder class, the User might be prompted to input their prescription.  The
system them parses the User's input to detect matches with both parsed and learned
predicates.  The components of any matching predicates are mapped to the parameters
defined in the \texttt{PredicateParamMappings} field. For example, our built-in parsed
predicates currently identify the frequency, time, event duration, start date, and the
duration of the reminder sequence (how long the reminder should remain on the
calendar). If these predicates fire due to the user's input, their slots are then
assigned to the corresponding script parameters, which can then be referenced in
subsequent turns; the script can populate a confirmation question (``You would like
your reminder at 7:00 a.m. Is that correct (Y/N)?'') without requiring the user to
first enter that value (``Please tell me what time you would like your reminder''), which
must be done if the slot is not filled.

The \textbf{TestPredicateTurn} has the following properties:

\FloatBarrier
\begin{table}[h!]
\centering
\begin{mdframed}
\begin{tabularx}{\textwidth}{l|l|l|X}
  Property & Type & Required & Description \\
  \hline
  \hline
  Question & string & yes & Prompts the user to enter a sentence describing the event. \\
  \hline
  Param & string & yes & The name of the parameter that will store the answer to the
  \texttt{Question}. \\
  \hline
  CountParam & string & yes & The name of the parameter that will store the number of
  phrases entered by the user (separated by "then''; see the discussion in the example). \\
  \hline
  PredicateParamMappings & struct & yes & A structure containing the names of the parsed
  predicates that are applicable to this application, and for each, the name of the
  parameter that will store the value of that parsed predicate. See examples below. \\
  \hline
  PredicateCountParam & NUMBER & yes & The name of the parameter that will receive
  the number of parsed predicates found. \\
\end{tabularx}
\end{mdframed}
\caption{TestPredicateTurn properties.}
\end{table}
\FloatBarrier

\noindent \textbf{Example:}

\begin{minipage}{\linewidth}
\begin{quote}
\begin{verbatim}
TestPredicateTurn:
    Question: Please read me the instructions on your prescription.
    Param: description_string
    CountParam: count
    PredicateParamMappings:
        Pred_Freq: frequency
        Pred_SequenceDuration: sequence_duration
        Pred_EventDuration: event_duration
        Pred_StartDate: start_date
    PredicateCountParam: description_predicate_count
\end{verbatim}
\end{quote}
\end{minipage}
\bigskip

In this example, the user will input a sentence in response to the \texttt{Question}.
First, the system splits the sentence into phrases by splitting on the "then'' keyword;
this can be extended to other keywords or syntactic structures. The number of clauses
found is stored in the parameter named by the \texttt{CountParam} property.
The system then parses each clause to extract parsed predicates from it.

The system creates a set of parameters for each clause, regardless of whether the
phrase contained a parsed predicate or not. For each clause, numbered from 0 to
the value in the \texttt{CountParam} parameter, the system creates a parameter for
each parsed predicate, with the ordinal of that clause appended to the
parameter name specified in the \textbf{TestPredicateTurn}. In the above example,
if the user's sentence contained 3 clauses, the system would create the following 
parameters \texttt{frequency.0}, \texttt{frequency.1}, \texttt{frequency.2}, and similar
parameters for the other parsed predicates found in the clause. These parameters are used
by a subsequent \nameref{turn:NIterations} turn to engage in a dialog with the User for
each clause.

FDL currently supports the built-in parsed predicate types in table \ref{tab:BIPPs}.

\FloatBarrier
\begin{table}[h!]
\centering
\begin{mdframed}
\begin{tabularx}{\textwidth}{X|l|X}
  Name & PropertyName & Description \\
  \hline
  \hline
  Pred\_Freq \newline ``\texttt{frequency}'' &  & A frequency such as "every day",
  "daily", "twice a day", "2-3 times a day", etc. If there is a single value ("twice a
  day") then an IntervalValue is created; otherwise ("2-3 times a day") an
  IntervalValueRange is created, and the User will be prompted to select a single value
  in that range. Pred\_Freq has the following three properties. \\
  \cline{2-3}
   & period & The period of the frequency, e.g. daily. \\
   & number\_of\_events\_per\_period & E.g., for "twice daily" this would be two. \\
   & start\_time & The starting time for each period; for "daily'' this might be "7 am''. \\
  \hline
  Pred\_SequenceDuration \newline ``\texttt{sequence\_duration}'' &  & The duration of
  the entire sequence; for the Event Reminder, this is how long the reminder should remain
  on the calendar. \\
  \hline
  Pred\_EventDuration \newline ``\texttt{event\_duration}'' &  & The duration of a single
  event; for the Event Reminder, this might be how long a single appointment would take. \\
  \hline
  Pred\_StartDate \newline ``\texttt{start\_date}' &  & E.g. for the Event Reminder, this
  is the date of the first reminder. \\
\end{tabularx}
\end{mdframed}
\caption{Built-In Parsed Predicates; the \texttt{Name} column shows, in quotes, the name
of the parameter to be given the predicate value in the Medication Reminder example, and
the \texttt{PropertyName} column shows the names the FDL system assigns to these
properties.}
\label{tab:BIPPs}
\end{table}
\FloatBarrier

Once the \texttt{TestPredicateTurn} is complete, a subsequent series of \texttt{Turn}s, usually
within an \texttt{NIterations Turn}, presents confirmation of each parsed predicate that
was detected in the clause, as well as asking for any that were not (and any other values
needed). If the User's sentence fully specifies all the necessary information (including
all required parsed predicates), then a single confirmation sentence can be presented;
in this case it would be necessary to confirm each individual value only if the User rejects this
initial confirmation.

\subsubsection{NIterations}
\label{turn:NIterations}

An \texttt{NIterations} composite turn is similar to a \texttt{for-loop} in most programming
languages. It has the following properties:

The \textbf{NIterations Turn} has the following properties:

\FloatBarrier
\begin{table}[h!]
\centering
\begin{mdframed}
\begin{tabularx}{\textwidth}{l|l|l|X}
  Property & Type & Required & Description \\
  \hline
  \hline
  IterVar & string & yes & The name of the parameter holding iteration variable (the
  "i" in the usual for loop, zero-based). \\
  \hline
  N & string & yes & The name of the parameter holding the maximum number of iterations. \\
  \hline
  Turns & Turn list & yes & The Turns to be executed on each iteration (the body of the
  for-loop). \\
\end{tabularx}
\end{mdframed}
\caption{NIterations properties.}
\end{table}
\FloatBarrier

\noindent \textbf{Example:}

\begin{minipage}{\linewidth}
\begin{quote}
\begin{verbatim}
	NIterations:
	  N: '10'
	  IterVar: i
	  Turns:
	    - Prompt: Now ${i}!
\end{verbatim}
\end{quote}
\end{minipage}

is similar to

\begin{minipage}{\linewidth}
\begin{quote}
\begin{verbatim}
	for (int i = 0; i <= 10; i++) {
	    printf "Now \%d!" i
	}
\end{verbatim}
\end{quote}
\end{minipage}
\bigskip

To present the iteration number in a more user-friendly way, the system provides a 
special parameter \texttt{\$\{ordinal\}} that maps the \texttt{IterVar} value (which is
zero-based) to the corresponding ordinal word (which is one-based). For example, in
iteration 0 (i.e. "i = 0"), then \texttt{\$\{ordinal\}} has a value of "first".

\texttt{NIterations} loops can be nested; the \texttt{Turns} list may contain an inner 
\texttt{NIterations} turn. In this case, the \texttt{\$\{ordinal\}} special parameter is
scoped to the inner iteration's \texttt{IterVar}.

It is often possible to set the parameter reference \texttt{N} to a previously assigned
NumberValue. For example, in the prevous example, we define a parameter \texttt{count}
as the number of clauses in the sentence; to iterate over these clauses, the \texttt{N}
parameter would be set to \texttt{\$\{count\}}.

\noindent \textbf{Example:}

\begin{minipage}{\linewidth}
\begin{quote}
\begin{verbatim}
NIterations:
	  N: ${count}
	  IterVar: i
	  Turns:
	    - Prompt: Let's confirm the ${ordinal} reminder.
	    - ParamValueTurn:
	        Question: Please tell me over what duration you'd like your reminders set.
	        Param: duration.${i}
	        Type: INTERVAL
\end{verbatim}
\end{quote}
\end{minipage}
\bigskip

The \texttt{NIterations} step allows date-sequencing logic. For the first iteration,
the frequency start\_date is as set by the user. In subsequent iterations, the start\_date
is the end date of the previous iteration, and this can be tracked in a script parameter
using the \texttt{SetParamValues} turn.

\noindent \textbf{Example:}

\begin{minipage}{\linewidth}
\begin{quote}
\begin{verbatim}
	SetParamValues:
	    Param: next_start_date
	    Type: DATE
	    Value: ${start_date.${i} + sequence_duration.${i}}
\end{verbatim}
\end{quote}
\end{minipage}
\bigskip

\subsubsection{SetParamValues}
\label{turn:SetParamValues}

An \texttt{SetParamValues} turn allows the Designer to set parameter values directly
in the script. 

The \textbf{SetParamValues Turn} has a list of structures that specify the parameters
to be set; each has the following properties:

\FloatBarrier
\begin{table}[h!]
\centering
\begin{mdframed}
\begin{tabularx}{\textwidth}{l|l|l|X}
  Property & Type & Required & Description \\
  \hline
  \hline
  Param & string & yes & The name of the parameter that will receive the value. \\
  \hline
  Type & ParamType & yes & The type of the parameter, from \fullref{sec:ParameterTypes}. \\
  \hline
  Value & string & yes & The value to be set, converted to \texttt{Type}. \\
\end{tabularx}
\end{mdframed}
\caption{SetParamValues properties.}
\end{table}
\FloatBarrier

\noindent \textbf{Example:}

\begin{minipage}{\linewidth}
\begin{quote}
\begin{verbatim}
	SetParamValues:
	    Param: i
	    Type: NUMBER
	    Value: 0
\end{verbatim}
\end{quote}
\end{minipage}
\bigskip

\subsubsection{RemoveParamValues}
\label{turn:RemoveParamValues}

An \texttt{RemoveParamValues} turn allows the Designer to remove parameter values directly
from the script. 

The \textbf{RemoveParamValues Turn} has a list of the following property:

\FloatBarrier
\begin{table}[h!]
\centering
\begin{mdframed}
\begin{tabularx}{\textwidth}{l|l|l|X}
  Property & Type & Required & Description \\
  \hline
  \hline
  Param & string & yes & The name of the parameter that will be removed. \\
\end{tabularx}
\end{mdframed}
\caption{RemoveParamValues properties.}
\end{table}
\FloatBarrier

\noindent \textbf{Example:}

\begin{minipage}{\linewidth}
\begin{quote}
\begin{verbatim}
	RemoveParamValues:
	    Param: frequency.${i}.period
	    Param: i
\end{verbatim}
\end{quote}
\end{minipage}
\bigskip

\subsubsection{ScriptConditionalActions}
\label{turn:ScriptConditionalActions}

A \texttt{ScriptConditionalActions} is equivalent to an if-else-then dialog flow,
based on the boolean value of a condition written in a simple conditional grammar that
allows testing whether a script parameter is set or has an infinite value, comparing numeric
values (=, >, >=, etc.), and parentheses for precedence.

The \textbf{ScriptConditionalActions Turn} has a list of the following property:

\FloatBarrier
\begin{table}[h!]
\centering
\begin{mdframed}
\begin{tabularx}{\textwidth}{l|l|l|X}
  Property & Type & Required & Description \\
  \hline
  \hline
  Condition & string & yes & The condition to evaluate; see \fullref{sec:ConditionalParams}. \\
  \hline
  IfYesTurns & Turn list & yes & The turns to execute if \texttt{Condition} is true; may be a
  \nameref{turn:NoAction} turn. \\
  \hline
  IfNoTurns & Turn list & yes & The turns to execute if \texttt{Condition} is false; may be a
  \nameref{turn:NoAction} turn. \\
\end{tabularx}
\end{mdframed}
\caption{ScriptConditionalActions properties.}
\end{table}
\FloatBarrier

\noindent \textbf{Example:}

\begin{minipage}{\linewidth}
\begin{quote}
\begin{verbatim}
	ScriptConditionalActions:
	  Condition: ${i} > 0 AND start_date.${i} IS NOT SET AND next_start_date IS SET
	  IfYesTurns:
	    - SetParamValues:
	        - Param: start_date.${i}
	          Type: DATE
	          Value: ${next_start_date}
	  IfNoTurns:
	    - NoAction
\end{verbatim}
\end{quote}
\end{minipage}
\bigskip

See also \nameref{turn:EndIterationAction} and \nameref{turn:EndScriptAction}.

\subsubsection{UserConditionalActions}
\label{turn:UserConditionalActions}

An \texttt{UserConditionalActions} turn allows the Designer to remove parameter values directly
from the script. 

The \textbf{UserConditionalActions Turn} is similar to \texttt{ScriptConditionalActions},
but the condition is the boolean result of a question presented to the user. For example,
in the Reminder application, if all necessary slots are filled, the user will be presented
with a single sentence to confirm; if No is returned, then the script presents the series
of ParamValueTurns.

\FloatBarrier
\begin{table}[h!]
\centering
\begin{mdframed}
\begin{tabularx}{\textwidth}{l|l|l|X}
  Property & Type & Required & Description \\
  \hline
  \hline
  Question & string & yes & The question to ask the user; the user will be allowed to enter
  only a yes or no answer. \\
  \hline
  IfYesTurns & Turn list & yes & The turns to execute if the user answers "Yes"; may be a
  \nameref{turn:NoAction} turn. \\
  \hline
  IfNoTurns & Turn list & yes & The turns to execute if the user answers "No"; may be a
  \nameref{turn:NoAction} turn. \\
\end{tabularx}
\end{mdframed}
\caption{UserConditionalActions properties.}
\end{table}
\FloatBarrier

The FDL system will append ``(Y/N)?'' to the the \texttt{Question} when it displays it to
the Teacher or User.
\bigskip

\noindent \textbf{Example:}

\begin{minipage}{\linewidth}
\begin{quote}
\begin{verbatim}
	UserConditionalActions:
	  Condition: ${i} > 0 AND start_date.${i} IS NOT SET AND next_start_date IS SET
	  IfYesTurns:
	    - SetParamValues:
	        - Param: start_date.${i}
	          Type: DATE
	          Value: ${next_start_date}
	  IfNoTurns:
	    - NoAction
\end{verbatim}
\end{quote}
\end{minipage}
\bigskip

See also \nameref{turn:EndIterationAction} and \nameref{turn:EndScriptAction}.

\subsubsection{ParamValueOrConstantTurn}
\label{turn:ParamValueOrConstantTurn}

The \texttt{ParamValueOrConstantTurn} is similar to the \texttt{ParamValueTurn},
but also allows the Teacher to enter a constant value if so doing is more appropriate
for the app they  are creating.  For example, a Teacher creating a birthday reminder app
would not want to have the app ask the user how often the reminder should fire.

The system will ask the Teacher to select one of three options:
\begin{enumerate}
  \item \textit{Ask the user a question}: This will display the question that the user will be asked.
  If this option is selected, no confirmation is requested.
  \item \textit{Ask the user a different question}: This will enter into a dialog with the Teacher
  to specify the question to be asked.
  \item \textit{Specify a constant value}: This will enter into a dialog with the Teacher to specify
  the constant value to be used instead of asking the user a question.
\end{enumerate}

The properties of the \texttt{ParamValueOrConstantTurn} are:

\FloatBarrier
\begin{table}[h!]
\centering
\begin{mdframed}
\begin{tabularx}{\textwidth}{l|l|l|X}
  Property & Type & Required & Description \\
  \hline
  \hline
  Description & string & no & Provides a description that is shown only once per execution
  of the step \\
  \hline
  AskQuestionConfirm& string & yes & The text that will be shown for the first option above;
  it presents the question that the User would be asked.  \\
  \hline
  AskDifferentQuestion& string & yes & The text that will be shown for the second option above,
  allowing the Teacher to specify a different question. \\
  \hline
  AskSpecifyConstant& string & yes & The text that will be shown for the third option above,
  allowing the Teacher to specify a constant value. \\
  \hline
  QuestionLoopback& string & yes & The sentence the system will use to ask the Teacher to
  enter the new question to ask the User. \\
  \hline
  QuestionConfirm& string & yes & The sentence the system will use to ask the Teacher to
  confirm the new question to ask the User. \\
  \hline
  QuestionParam& string & yes & The name of the parameter that will store the question to
  ask the User. \\
  \hline
  QuestionDefault& string & yes & The default question to ask the User. \\
  \hline
  ConstantLoopback& string & yes & The sentence the system will use to ask the Teacher to
  enter the constant value. \\
  \hline
  ConstantConfirm& string & yes & The sentence the system will use to ask the Teacher to
  confirm the constant value. \\
  \hline
  ConstantParam& string & yes & The name of the parameter that will store the constant
  value. \\
  \hline
  ParamType & ParamType & yes & The type of the parameter, from
  \fullref{sec:ParameterTypes}. \\
\end{tabularx}
\end{mdframed}
\caption{UserConditionalActions properties.}
\end{table}
\FloatBarrier

The FDL system will append ``(Y/N)?'' to the the \texttt{QuestionConfirm} and
\texttt{ConstantConfirm} when it displays them to the Teacher.
\bigskip

\noindent \textbf{Example:}

\begin{minipage}{\linewidth}
\begin{quote}
\begin{verbatim}
	- ParamValueOrConstantTurn:
	    Description: I will ask the user how long to keep the reminder in their calendar.
	    AskQuestionConfirm: I will ask the user: "${ask_sequence_duration_language}".
	    AskDifferentQuestion: Ask the user a different question.
	    AskSpecifyConstant: Do not ask the User; instead, specify a constant value.
	    QuestionLoopback: Please enter the new question to ask the user.
	    QuestionConfirm: You want me to ask the user "${ask_sequence_duration_language}".
				${_IsThatCorrect}
	    QuestionParam: ask_sequence_duration_language
	    QuestionDefault: Please tell me how long to keep the reminder in your calendar.
	    ConstantLoopback: Please enter the constant value you would like to use here.
	    ConstantConfirm: You would like to stop the reminders after a constant period of
				${constant_sequence_duration}. ${_IsThatCorrect}
	    ConstantParam: constant_sequence_duration
	    ParamType: INTERVAL
\end{verbatim}
\end{quote}
\end{minipage}
\bigskip

\subsubsection{SetConstantValues}
\label{turn:SetConstantValues}

The \textbf{SetConstantValues Turn} sets script variables to constant values (such as those
set by \texttt{ParamValueOrConstantTurn}) if they were not already set (such as by a
\texttt{TestPredicateTurn}). 

The \textbf{SetConstantValues Turn} has a list of structures that specify the parameters
to be set; each has the following properties:

\FloatBarrier
\begin{table}[h!]
\centering
\begin{mdframed}
\begin{tabularx}{\textwidth}{l|l|l|X}
  Property & Type & Required & Description \\
  \hline
  \hline
  ConstantParam & string & yes & The name of the parameter that contains the constant
  value. \\
  \hline
  Param & string & yes & The name of the parameter to receive the constant value. \\
\end{tabularx}
\end{mdframed}
\caption{SetConstantValues properties.}
\end{table}
\FloatBarrier

\noindent \textbf{Example:}

\begin{minipage}{\linewidth}
\begin{quote}
\begin{verbatim}
	- SetConstantValues:
	    - ConstantParam: constant_sequence_duration
	      Param: sequence_duration.${i}
	    - ConstantParam: constant_period
	      Param: frequency.${i}.period
\end{verbatim}
\end{quote}
\end{minipage}
\bigskip

\subsubsection{EndIterationAction}
\label{turn:EndIterationAction}

The \textbf{EndIterationAction Turn} exits an \texttt{NIterations} loop. It contains the
following property:

\FloatBarrier
\begin{table}[h!]
\centering
\begin{mdframed}
\begin{tabularx}{\textwidth}{l|l|l|X}
  Property & Type & Required & Description \\
  \hline
  \hline
  CurrentIterationOnly & bool & yes & If true, exits only the innermost iteration;
  otherwise, exits all loops. \\
\end{tabularx}
\end{mdframed}
\caption{EndIterationsAction properties.}
\end{table}
\FloatBarrier

\noindent \textbf{Example:}

\begin{minipage}{\linewidth}
\begin{quote}
\begin{verbatim}
	- EndIterationAction:
	    CurrentIterationOnly: true
\end{verbatim}
\end{quote}
\end{minipage}
\bigskip

\subsubsection{EndScriptAction}
\label{turn:EndScriptAction}

The \textbf{EndScriptAction Turn} exits the script. It contains the following property:

\FloatBarrier
\begin{table}[h!]
\centering
\begin{mdframed}
\begin{tabularx}{\textwidth}{l|l|l|X}
  Property & Type & Required & Description \\
  \hline
  \hline
  IsSuccess & bool & yes & If true, exits successfully; otherwise exits with an error. \\
\end{tabularx}
\end{mdframed}
\caption{EndScriptAction properties.}
\end{table}
\FloatBarrier

\noindent \textbf{Example:}

\begin{minipage}{\linewidth}
\begin{quote}
\begin{verbatim}
	- EndScriptAction:
	    IsSuccess: true
\end{verbatim}
\end{quote}
\end{minipage}
\bigskip

\subsubsection{NoAction}
\label{turn:NoAction}

The \textbf{NoAction Turn} is a no-op; it is an empty branch in the conditional actions
turns and has no parameters.

\subsection{Finding the App the User Wants}
\label{sec:FindingTheApp}

When the user wants to use an app to perform an action, the FDL system will ask for a
sentence that describes the action, and will then parse this action and find any matching
\texttt{LearnedPredicates} that have a \texttt{Handler} (see \nameref{sec:Handlers})
for that action. Following are the steps taken by the FDL system when the user wants
to use an app to perform a \texttt{create} action; for example, using the Medication
Reminder app to create a reminder to take medication.
\begin{enumerate}
	\item The FDL system asks the user to enter a sentence describing the activity to create
a reminder for. It parses this sentence, determines which templates match it, and forms a
set of all the \texttt{LearnedPredicate}s that are fired by those templates and have an
entry in their Handlers set for the ActionName \texttt{create}.
	\begin{enumerate}
		\item If there is only one such \texttt{LearnedPredicate} and it has only one
\texttt{Handler} for the \texttt{create} ActionName, then that app is selected
automatically.
		\item If there is no matching \texttt{LearnedPredicate}, the FDL system presents a list
of all available apps for which one or more \texttt{LearnedPredicates} have a
\texttt{Handler} with the \texttt{create} ActionName. These apps are identified by their
scripts' \texttt{\_\_ACTION\_DESCRIPTION\_\_}s. The user is asked to select an app.
		\item If there is more than one matching \texttt{LearnedPredicate} or one or more
matching \texttt{LearnedPredicate}s have more than one \texttt{Handler} with the
\texttt{create} ActionName, the FDL system presents a list of all matching apps, again
identified by their scripts' \texttt{\_\_ACTION\_DESCRIPTION\_\_}s. The user is asked
to select an app.
\end{enumerate}
\end{enumerate}

When the user selects an app (or if an app is selected automatically because it is the
only matching app), then the FDL system asks the user to confirm the action by presenting
the app's \texttt{\_\_ACTION\_CONFIRMATION\_\_} question. If the user says ``Yes'', then
the FDL system loads the script for that app (as identified by the selected
\texttt{Handler}'s ScriptName) and begins the app's dialog with the user.

%% file: appendix_2.tex
\newpage
\section{Appendix 2: TAL's American Slang and Abbreviations Map}
\label{apx:slangMap}
During preprocessing, tokens found in the left column in user input data are mapped to the
right (see Section \ref{sec:talPreprocessing}).  Teachers can edit this file if they wish:
note that here, a teacher added the mapping ``meds'' to ``medications'' for their app
(this was not necessary, as otherwise TAL would ask the teacher what ``meds'' means the
first time the teacher used it, and add it as a new definition to its taxonomy).

\begin{table}[!ht]
\centering
\begin{tabular}{cccccccc}
\multicolumn{2}{c}{ } & & \multicolumn{2}{c}{ } & & \multicolumn{2}{c}{ } \\
\cmidrule(r){1-2}
\cmidrule(lr){4-5}
\cmidrule(l){7-8}
wanna	&	want to && gonna	&	going to	&& shoulda	&	should have \\
coulda	&	could have	&& woulda	&	would have	&&	sorta	&	somewhat \\
kinda	&	somewhat	&& tonite	&	tonight	&& pls	&	please \\
thx	&	thanks	&& meds	&	medications	&& wont	&	will not \\
cant	&	can not	&& aren't	&	are not	&&	can't	&	can not \\
didn't	&	did not	&& don't	&	do not	&&	doesn't	&	does not \\
hadn't	&	had not	&& hasn't	&	has not	&&	haven't	&	have not \\
he'd	&	he would	&& he'll	&	he will	&&	he's	&	he is \\
i'd	&	i would	&& i'll	&	i will	&&	i'm	&	i am \\
isn't	&	is not	&& it'd	&	it would	&&	it's	&	it is \\
it'll	&	it will	&& i've	&	i have	&&	let's	&	let us \\
mustn't	&	must not	&& needn't	&	need not	&&	she'd	&	she would \\
she'll	&	she will	&& she's	&	she is	&&	shouldn't	&	should not \\
that'd	&	that would	&& that'll	&	that will	&&	that's	&	that is \\
there'll	&	there will	&& there's	&	there is	&&	there've	&	there have \\
they'd	&	they would	&& they'll	&	they will	&& they're	&	they are \\
they've	&	they have	&& wasn't	&	was not	&&	we'll	&	we will \\
we'd	&	we would	&& we're	&	we are	&&	we've	&	we have \\
weren't	&	were not	&& won't	&	will not	&&	wouldn't	&	would not \\
you'd	&	you would	&& you'll	&	you will	&&	you're	&	you are \\
you've	&	you have &&	&	&& &  \\
\cmidrule(r){1-2}
\cmidrule(lr){4-5}
\cmidrule(l){7-8}
\end{tabular}
\caption{TAL's slang and abbreviations map.}
\label{tab:slangMap}
\end{table}

%% file: main.bbl
\begin{thebibliography}{10}

\bibitem{NumberOfAndroidApps:2016}
AppBrain.
\newblock Number of available apps, 2016.
\newblock \url{http://www.appbrain.com/stats/number-of-android-apps}.

\bibitem{Boehm:1988}
B.W. Boehm.
\newblock A spiral model of software development and enhancement.
\newblock {\em Computer}, 21(5):61--72, 1988.

\bibitem{Burges:2013}
C.J.C. Burges.
\newblock Towards the machine comprehension of text: An essay.
\newblock Technical Report MSR-TR-2013-125, Microsoft Research, 2013.

\bibitem{Crevier:1993}
D.~Crevier.
\newblock {\em AI: The tumultuous history of the search for artificial
  intelligence}.
\newblock Basic Books, Inc., 1993.

\bibitem{Fellbaum:1998}
C.~Fellbaum.
\newblock {\em WordNet}.
\newblock Wiley Online Library, 1998.

\bibitem{GooBenCou:2016}
I.~Goodfellow, Y.~Bengio, and A.~Courville.
\newblock Deep learning.
\newblock Book in preparation for MIT Press:
  \url{http://www.deeplearningbook.org}, 2016.

\bibitem{Hixon:2015}
B.~Hixon, P.~Clark, and H.~Hajishirzi.
\newblock Learning knowledge graphs for question answering through
  conversational dialog.
\newblock In {\em Proceedings of the the 2015 Conference of the North American
  Chapter of the Association for Computational Linguistics: Human Language
  Technologies, Denver, Colorado, USA}, 2015.

\bibitem{HuYanSalXin:2016}
Z.~Hu, X.~Ma, Z.~Liu, E.~Hovy, and E.~Xing.
\newblock Harnessing deep neural networks with logic rules.
\newblock In {\em 54th Annual Meeting of the Association for Computational
  Linguistics}, 2016.

\bibitem{KimBanLi:2016}
S.~Kim, R.E. Banchs, and H~Li.
\newblock Exploring convolutional and recurrent neural networks in sequential
  labelling for dialogue topic tracking.
\newblock In {\em 54th Annual Meeting of the Association for Computational
  Linguistics}, 2016.

\bibitem{LiGalBroGaoDol:2016}
J.~Li, M.~Galley, C.~Brockett, J.~Gao, and B.~Dolan.
\newblock A persona-based neural conversation model.
\newblock In {\em 54th Annual Meeting of the Association for Computational
  Linguistics}, 2016.

\bibitem{Liang:2016}
P.~Liang.
\newblock Learning executable semantic parsers for natural language
  understanding.
\newblock {\em Communications of the ACM}, 59, 2016.

\bibitem{Maes:1994}
Pattie Maes.
\newblock Agents that reduce work and information overload.
\newblock {\em Communications of the ACM}, 37(7):30--40, 1994.

\bibitem{ManSurBauFinBetMcc:2014}
C.D. Manning, M.~Surdeanu, J.~Bauer, J.~Finkel, S.J. Bethard, and D.~McClosky.
\newblock The {Stanford} {CoreNLP} natural language processing toolkit.
\newblock In {\em Association for Computational Linguistics (ACL) System
  Demonstrations}, pages 55--60, 2014.

\bibitem{Mueller:2014}
Erik~T Mueller.
\newblock {\em Commonsense Reasoning}.
\newblock Morgan Kaufmann, 2014.

\bibitem{Myers:2007}
Karen Myers, Pauline Berry, Jim Blythe, Ken Conley, Melinda Gervasio, Deborah~L
  McGuinness, David Morley, Avi Pfeffer, Martha Pollack, and Milind Tambe.
\newblock An intelligent personal assistant for task and time management.
\newblock {\em AI Magazine}, 28(2):47, 2007.

\bibitem{NguYosClu:2014}
A.~Nguyen, J.~Yosinski, and J.~Clune.
\newblock Deep neural networks are easily fooled: High confidence predictions
  for unrecognizable images.
\newblock In {\em 2015 IEEE Conference on Computer Vision and Pattern
  Recognition (CVPR)}, pages 427--436. IEEE, 2015.

\bibitem{PanYan:2010}
Sinno~Jialin Pan and Qiang Yang.
\newblock A survey on transfer learning.
\newblock {\em IEEE Transactions on knowledge and data engineering},
  22(10):1345--1359, 2010.

\bibitem{PicMoo:2016}
K.~Pichotta and R.J. Mooney.
\newblock Statistical script learning with recurrent neural networks.
\newblock In {\em Proceedings of the Workshop on Uphill Battles in Language
  Processing (UBLP) at EMNLP 2016}, 2016.

\bibitem{QuiChoGaoSuzTouGamYihVanChe12}
C.~Quirk, P.~Choudhury, J.~Gao, H.~Suzuki, K.~Toutanova, M.~Gamon, W.~Yih,
  L.~Vanderwende, and C.~Cherry.
\newblock Msr splat, a language analysis toolkit.
\newblock In {\em Proceedings of the NAACL-HLT 2012: Demonstration Session},
  pages 21--24, 2012.

\bibitem{RajZhaLopLia:2016}
P.~Rajpurkar, J.~Zhang, K.~Lopyrev, and P.~Liang.
\newblock Squad: 100,000+ questions for machine comprehension of text.
\newblock In {\em Empirical Methods in Natural Language Processing (EMNLP)},
  2016.

\bibitem{RicBurRen:2013}
M.~Richardson, C.J.C. Burges, and E.~Renshaw.
\newblock Mctest: A challenge dataset for the open-domain machine comprehension
  of text.
\newblock In {\em Empirical Methods in Natural Language Processing (EMNLP)},
  2013.

\bibitem{RicDom:2006}
M.~Richardson and P.~Domingos.
\newblock Markov logic networks.
\newblock {\em Machine learning}, 62(1-2):107--136, 2006.

\bibitem{Schank:1977}
Roger~C Schank and Robert~P Abelson.
\newblock {\em Scripts, plans, goals, and understanding: An inquiry into human
  knowledge structures}.
\newblock Psychology Press, 1977.

\bibitem{SimChiLakChaBotSuaGanAmeVerSuh:2014}
P.~Simard, D.~Chickering, A.~Lakshmiratan, D.~Charles, L.~Bottou, C.~Suarez,
  D.~Grangier, S.~Amershi, J.~Verwey, and J.~Suh.
\newblock Ice: enabling non-experts to build models interactively for
  large-scale lopsided problems.
\newblock {\em arXiv preprint arXiv:1409.4814}, 2014.

\bibitem{SzeZarSutBruErhGooFer:2013}
C.~Szegedy, W.~Zaremba, I.~Sutskever, J.~Bruna, D.~Erhan, I.~Goodfellow, and
  R.~Fergus.
\newblock Intriguing properties of neural networks.
\newblock {\em arXiv preprint arXiv:1312.6199}, 2013.

\bibitem{TerTenGer:2016}
D.G.R. Tervo, J.B. Tenenbaum, and S.J. Gershman.
\newblock Toward the neural implememtaion of structure learning.
\newblock {\em Current Opinion in Neurobiology}, 37, 2016.

\bibitem{TriYeYuaHeBacSul:2016}
A.~Trischler, Z.~Ye, X.~Yuan, J.~He, P.~Bachman, and K.~Suleman.
\newblock A parallel-hierarchical model for machine comprehension on sparse
  data.
\newblock {\em arXiv preprint arXiv:1603.08884}, 2016.

\bibitem{Tur:2011}
G.~Tur and R.~De~Mori.
\newblock {\em Spoken language understanding: Systems for extracting semantic
  information from speech}.
\newblock John Wiley \& Sons, 2011.

\bibitem{WanBanGimMca:2015}
H.~Wang, M.~Bansal, K.~Gimpel, and D.~McAllester.
\newblock Machine comprehension with syntax, frames, and semantics.
\newblock {\em Volume 2: Short Papers}, page 700, 2015.

\bibitem{WanBenCol:2012}
L.~Wang, P.N. Bennett, and K.~Collins-Thompson.
\newblock Robust ranking models via risk-sensitive optimization.
\newblock In {\em Proceedings of the 35th international ACM SIGIR conference on
  Research and development in information retrieval}, pages 761--770. ACM,
  2012.

\bibitem{AIWinter:2016}
Wikipedia.
\newblock {AI} {W}inter, 2016.
\newblock \url{https://en.wikipedia.org/wiki/AI_winter}.

\bibitem{PredicateGrammar:2016}
Wikipedia.
\newblock Predicate (grammar), 2016.
\newblock \url{https://en.wikipedia.org/wiki/Predicate_(grammar)}.

\bibitem{WordNetForOtherLanguages:2016}
Wikipedia.
\newblock Wordnet for other languages, 2016.
\newblock \url{https://en.wikipedia.org/wiki/WordNet#Other_languages}.

\bibitem{WilKamMokMilZwe:2015}
J.D. Williams, E.~Kamal, H.A. Mokhtar~Ashour, J.~Miller, and G.~Zweig.
\newblock Fast and easy language understanding for dialog systems with
  microsoft language understanding intelligent service (luis).
\newblock In {\em 16th Annual Meeting of the Special Interest Group on
  Discourse and Dialogue}, page 159, 2015.

\bibitem{WilNirDasLakSuaRedZwe:2015}
J.D. Williams, N.B. Niraula, P.~Dasigi, A.~Lakshmiratan, C.~Suarez, M.~Reddy,
  and G.~Zweig.
\newblock Rapidly scaling dialog systems with interactive learning.
\newblock In {\em Natural Language Dialog Systems and Intelligent Assistants},
  pages 1--13. Springer, 2015.

\end{thebibliography}
